\documentclass[a4paper,fleqn]{cas-dc}

\usepackage[authoryear,longnamesfirst]{natbib}
\usepackage{framed,multirow}

\usepackage{amssymb}
\usepackage{latexsym}

\usepackage{url}
\usepackage{xcolor}
\definecolor{newcolor}{rgb}{.8,.349,.1}
\usepackage{graphicx}
\usepackage{algorithm2e}
\usepackage{algorithmicx}
\usepackage{amsmath}  
\usepackage{float}
\usepackage{booktabs}
\usepackage{listings}
\usepackage{lineno}
\usepackage{hyperref}

\def\tsc#1{\csdef{#1}{\textsc{\lowercase{#1}}\xspace}}
\tsc{WGM}
\tsc{QE}

\begin{document}
\let\WriteBookmarks\relax
\def\floatpagepagefraction{1}
\def\textpagefraction{.001}

\shorttitle{T2VEval: Benchmark Dataset and Objective Evaluation Method for T2V-generated Videos}

\shortauthors{Qi et al.}  

\title[mode = title]{T2VEval: Benchmark Dataset and Objective Evaluation Method for T2V-generated Videos}  
\author[1]{Zelu Qi}
\ead{theoneqi2001@cuc.edu.cn}
\author[1]{Ping Shi*}
\cortext[cor1]{Corresponding author}
\author[1]{Shuqi Wang}
\author[1]{Chaoyang Zhang}
\author[1]{Fei Zhao}
\author[1]{Zefeng Ying}
\author[1]{Da Pan}
\author[2]{Xi Yang}
\author[2]{Zheqi He}
\author[2]{Teng Dai}
\address[1]{School of Information and Communication Engineering, Communication University of China, No.1 East Street, Dingfuzhuang, Chaoyang District, Beijing, China}
\address[2]{Beijing Academy of Artificial Intelligence, No.150 Chengfu Road, Haidian District, Beijing, China}

\begin{abstract}
Recent advances in text-to-video (T2V) technology, as demonstrated by models such as Runway Gen-3, Pika, Sora, and Kling, have significantly broadened the applicability and popularity of the technology. This progress has created a growing demand for accurate quality assessment metrics to evaluate the perceptual quality of T2V-generated videos and optimize video generation models. However, assessing the quality of text-to-video outputs remain challenging due to the presence of highly complex distortions, such as unnatural actions and phenomena that defy human cognition. To address these challenges, we constructed T2VEval-Bench, a multi-dimensional benchmark dataset for text-to-video quality evaluation, which contains 148 textual prompts and 1,783 videos generated by 13 T2V models. To ensure a comprehensive evaluation, we scored each video on four dimensions in the subjective experiment, which are overall impression, text-video consistency, realness, and technical quality. Based on T2VEval-Bench, we developed T2VEval, a multi-branch fusion scheme for T2V quality evaluation. T2VEval assesses videos across three branches: text-video consistency, realness, and technical quality. Using an attention-based fusion module, T2VEval effectively integrates features from each branch and predicts scores with the aid of a large language model. Additionally, we implemented a divide-and-conquer training strategy, enabling each branch to learn targeted knowledge while maintaining synergy with the others. Experimental results demonstrate that T2VEval achieves state-of-the-art performance across multiple metrics. 
\end{abstract}

\begin{keywords}
Text-to-Video \sep Quality Assessment \sep Evaluation benchmark \sep Objective Evaluation
\end{keywords}

\maketitle

\section{Introduction}
\begin{figure}[h]
    \centering
    \includegraphics[scale=0.33]{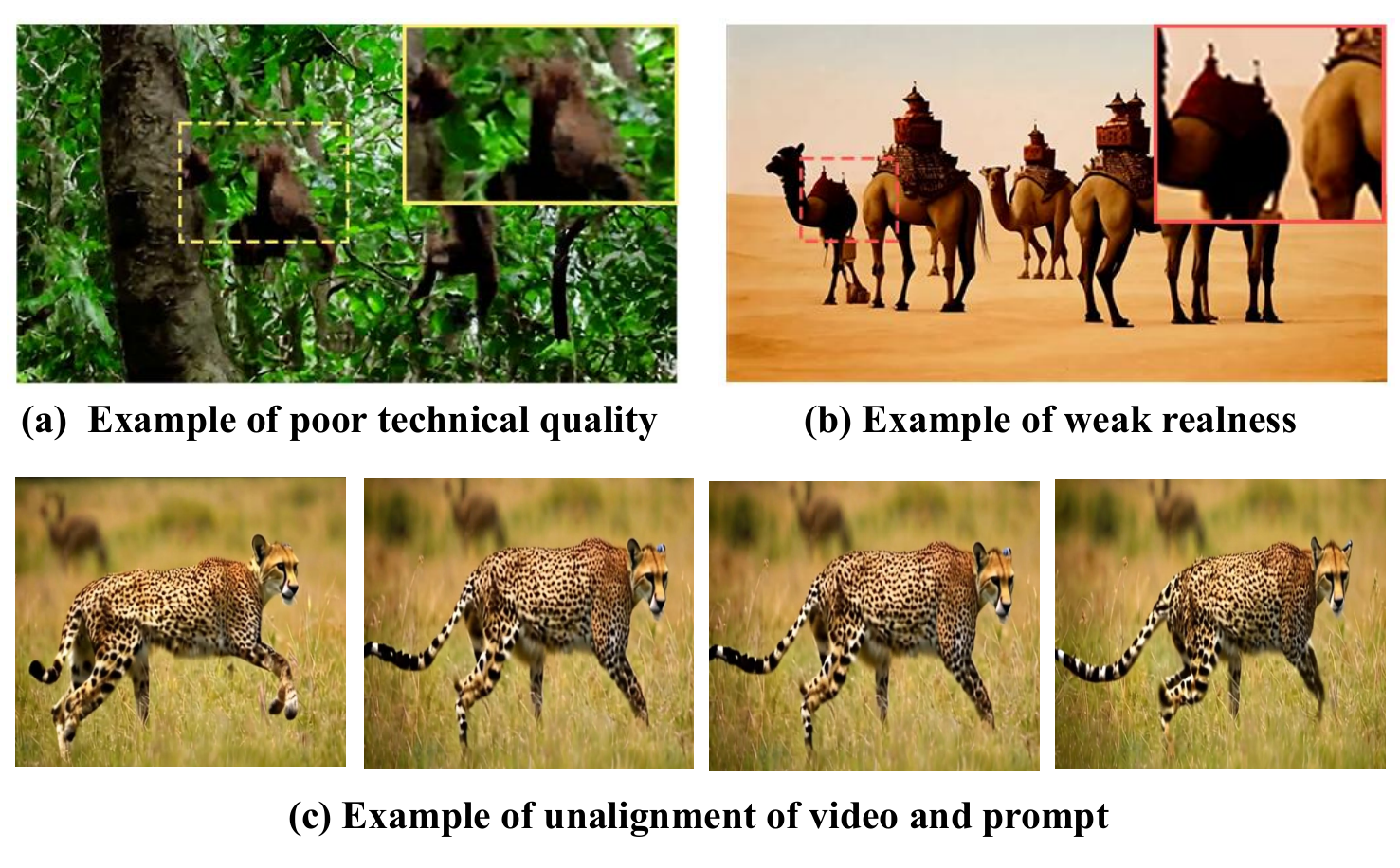}
    \caption{Examples of defects in T2V-generated videos. (a) As shown in the yellow matrix part of the picture, there is obvious distortion similar to "block effect" in the picture. (b) As shown in the red matrix part of the picture, the camel's limbs and head are "broken". (c) The prompt describes a cheetah chasing an antelope. However, the generated video only has a cheetah, no antelope, and the state of "chasing" is not simulated.}
    \label{fig1}
\end{figure}

Text-to-video (T2V) technology has made significant advancements in the last two years and attracted growing attention from both the research community and the general public. As an important branch of Generative Artificial Intelligence (GenAI), T2V models \cite{blattmann2023align,blattmann2023stable,ho2022imagen,singer2022make,wu2021godiva,wu2022nuwa,hong2022cogvideo,villegas2022phenaki,zhou2022magicvideo,khachatryan2023text2video,luo2023videofusion,he2022latent,wang2023modelscope,yang2024cogvideox,zhang2024show,wang2024recipe,videoworldsimulators2024} have developed rapidly recently. Contemporary T2V models, such as Sora, Runway Gen-3, PixVerse, Kling, etc. have enabled video generation from free-form textual prompts and demonstrated impressive visual fidelity and alignment with the prompts. However, as illustrated in Figure \ref{fig1}, the T2V-generated videos still exhibit several notable deficiencies that cannot be overlooked. These shortcomings can significantly detract from the user’s viewing experience. Therefore, developing a comprehensive and effective text-to-video quality assessment strategy is essential to ensure an optimal user experience and to guide the optimization of T2V models. 

\begin{figure*}[!bt]
\centering
\includegraphics[scale=0.6]{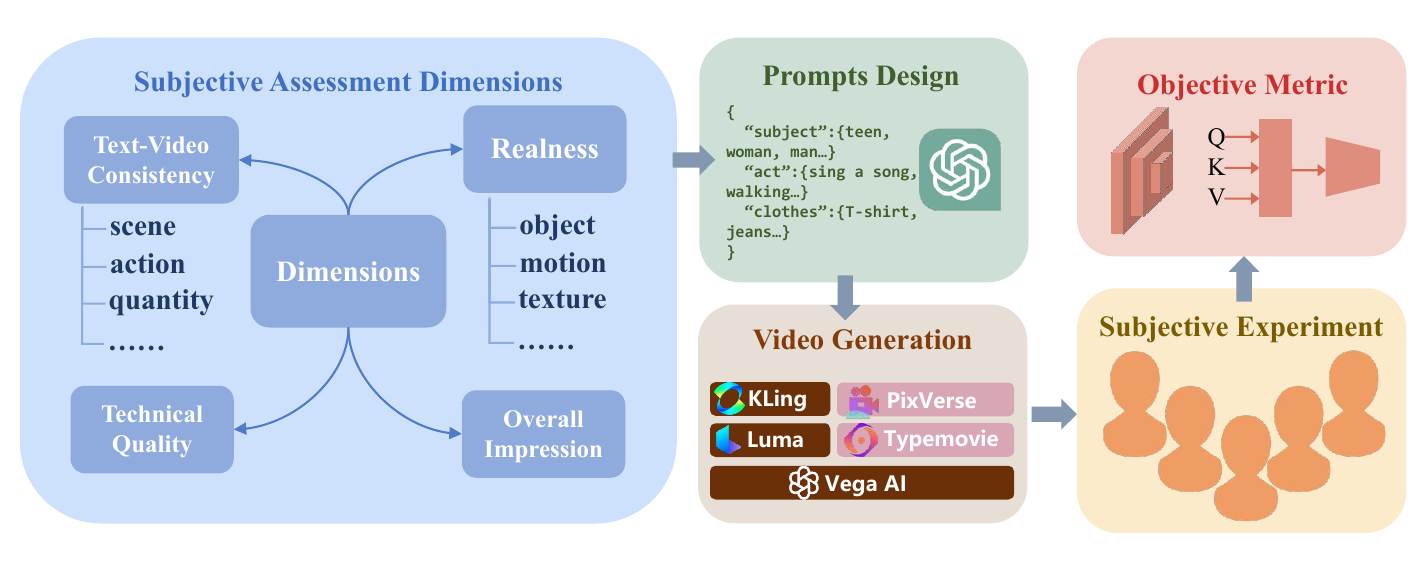}
\caption{The overall framework of our work, which includes the T2V evaluation benchmark T2VEval-Bench, encompassing standardized evaluation dimensions, textual prompt generation, video generation and subjective experiment. Building upon this benchmark, we further designed T2VEval, an objective evaluation method for T2V-generated videos.}
\label{fig2}
\end{figure*}

In recent years, numerous studies have focused on establishing T2V evaluation benchmarks and automated metrics \cite{liu2024fetv, liu2024evalcrafter,wu2024towards,huang2024vbench,zhang2024rethinking,kou2024subjective, sun2024t2v} for assessing the text-to-video quality across dimensions such as visual quality, temporal consistency, and text-video alignment. Although these approach offers a more detailed representation of the generative abilities of T2V models, it may lead to metrics that are less applicable and generalizable due to their over-specialization to specific evaluation dimensions. For instance, readability assessments of the text-related elements in T2V-generated videos become redundant in contexts where textual content is absence or not the primary focus. Furthermore, many works design automated metrics by directly employing models such as DOVER\cite{wu2023exploring} and RAFT\cite{teed2020raft}, which are trained and tested on datasets consisting of human-made videos rather than AI-generated videos. This operation ignores the inherent statistical distribution differences between these two classes of videos. This practice overlooking the inherent statistical distribution differences between real videos and text-generated videos. In addition, some studies, similar to traditional video and image quality assessments, use a single score to represent the quality of T2V-generated videos. Although this approach simplifies the evaluation process and enhances metric applicability, it can easily overlook the strengths and weaknesses of certain T2V models in specific generation tasks.

To overcome the aforementioned challenges,  we first constructed \textbf{T2VEval-Bench}, a multi-dimensional benchmark dataset for text-to-video quality evaluation. The construction process primarily consists of four parts: the division of subjective evaluation dimensions, the compilation and screening of textual prompts, video generation, and the collection of subjective scores. Specifically, we achieved a balanced trade-off between evaluation granularity and metric applicability by defining four evaluation dimensions, including \textbf{overall impression, text-video consistency, realness, and technical quality}, considering the prevalent defects in current text-to-video generation. Overall impression refers to the evaluators' general and intuitive impression of the video, which serve as a holistic measure of user satisfaction. For prompt generation, we utilized large language models (LLMs) in conjunction with manual verification to create 100 textual prompts covering four themes: \textbf{human, animal, landscape, and imagination}. Additionally, we designed prompts to reflect regional and cultural characteristics as well as physical laws. We also incorporated 48 publicly available prompts from Sora's demos. Subsequently, we employed 12 state-of-the-art T2V models to generate 1,735 videos\footnote{Some T2V models failed to generate corresponding videos for certain prompts.} and incorporated 48 publicly available videos from Sora, resulting in a total of 1,783 videos. Building on this dataset, we conducted extensive subjective evaluation experiments in a strictly controlled laboratory environment and collected the Mean Opinion Scores (MOS) on four evaluation dimensions.

Based on the T2VEval-Bench dataset, we further designed \textbf{T2VEval}, an objective evaluation method for T2V-generated. Previous works, such as FETV \cite{liu2024fetv}, EvalCrafter \cite{liu2024evalcrafter}, and VBench \cite{huang2024vbench}, have developed dedicated evaluation metrics for each evaluation dimension. However, these metrics generally lack broad applicability, and they directly employ models trained on datasets with human-made videos, thereby overlooking the inherent distributional differences between human-made videos and AI-generated videos. Other studies like T2VQA\cite{kou2024subjective} and AIGC-VQA \cite{lu2024aigc} have implemented end-to-end evaluations of videos, but they only consider video quality and text-video alignment. Therefore, we designed a multi-branch fusion evaluation scheme, which includes \textbf{a technical quality encoding branch, a video realness encoding branch, and a text-video alignment encoding branch}. The technical quality encoder is implemented based on the Swin-3D network \cite{liu2022video}, the video realness encoder utilizes the ConvLexNet-3D network \cite{liu2022convnet}, and the text-video alignment is achieved using the BLIP network \cite{li2022blip}. Considering that technical quality and realness rely solely on visual perception, we prioritize the fusion of quality and realness features before integrating them with text-video alignment features. Additionally, we adopted a progressive training strategy that ensures the independent effectiveness of each branch while promoting collaboration among the branches. Experimental results show that T2VEval effectively assesses the performance of different T2V-generated videos and has broad potential applications, achieved state-of-the-art performance on both T2VEval-Bench and T2VQA-DB\cite{kou2024subjective}.

\section{RELATED WORK}

\subsection{Subjective evaluation  and dataset for T2V Generation}

Chivileva et al. \cite{chivileva2023measuring} constructed the first T2V generated video evaluation dataset. They obtained 201 textual prompts by combining ChatGPT and manual writing, covering topics such as famous figures, regions, and special festivals. Five T2V models, namely Aphantasia, Text2Video-Zero, T2VSynthesis, Tune-a-Video, and Video Fusion, were selected to generate 1005 videos. Subsequently, 24 subjects were invited to subjectively score the generated videos from the perspectives of naturalness and consistency. Liu et al. proposed FETV \cite{liu2024fetv}. They designed a set of diverse and fine-grained textual prompts and classified them into three aspects: main content, control attributes, and text complexity. At the same time, a time category was specifically designed for text-to-video, covering spatial and temporal dimensions. On this basis, the authors selected four T2V models, including CogVideo, Text2Video-zero, ModelScopeT2V, and ZeroScope, and conducted subjective evaluations in terms of static quality, temporal quality, overall alignment, and fine-grained alignment. EvalCrafter \cite{liu2024evalcrafter} first constructed a comprehensive textual prompt corpus covering daily objects, attributes, and actions. It specified the metadata of common elements in the real world and expanded them using large language models like ChatGPT to obtain complete text prompts. It also enriched and expanded the prompts through those written by real-world users and from text-to-image (T2I) datasets. Subsequently, 2500 videos were generated on 5 T2V models, and subjective evaluations were carried out in terms of video quality, text-video consistency, motion quality, and time consistency. Wu et al. \cite{wu2024towards} constructed the TVGE dataset. They adopted the prompt set proposed in EvalCrafter \cite{liu2024evalcrafter} and generated 2543 videos on 5 T2V models. Ten professional scorers were invited to score the videos in terms of video quality and video-text consistency. Huang et al. \cite{huang2024vbench} believed that the existing evaluation benchmarks could not accurately reflect the highlights and problems of different generation models. Based on this, they proposed Vbench, which further refined the evaluation dimensions of T2V. Sixteen fine-grained evaluation dimensions for generated videos were designed, including subject identity inconsistency, motion smoothness, time flicker, and spatial relationship. One hundred text prompts were designed for each evaluation dimension. In addition, the authors also evaluated T2V models for different content categories, such as people, animals, and landscapes. A total of 8 content categories were considered, and 100 prompts were designed for each category. Based on this, the authors conducted subjective evaluations on 4 T2V models and collected corresponding 16-dimensional scores. Li et al. \cite{zhang2024benchmarking} constructed the LGVQ dataset, which contains 468 text prompts covering combinations of different foreground contents, background contents, and motion states. Six T2V models were used to generate 2808 videos, and 60 subjects were invited to perceptually score the spatial quality, temporal quality, and text-video alignment of each video. Based on the LGVQ dataset, the authors carried out benchmark tests using 14 IQA methods, 16 VQA methods, and 9 methods based on CLIP and visual question answering. Zhang et al. \cite{zhang2024rethinking} pointed out that the existing subjective evaluation benchmarks for T2V generally have limitations in reliability, reproducibility, and practicability. In view of this, this paper proposes the T2VHE benchmark, which specifies the evaluation dimensions (including video quality, temporal quality, motion quality, text alignment, etc.) used in T2V evaluation, the annotator training process and example descriptions, and a user-friendly GUI interface. The authors further proposed a dynamic scoring scheme. By combining the automatic scoring method to pre-annotate the videos, the amount of manually annotated data was subsequently reduced by 50\%. Kou et al. \cite{kou2024subjective} believed that the current T2V evaluation benchmarks have overly complex divisions of evaluation dimensions and lack a simple and effective evaluation index. Therefore, the authors constructed T2VQA-DB and invited 24 subjects to score 10,000 text-generated videos. The subjects were required to give an overall score considering video quality and text-video consistency.  Considering that current AIGVs usually have unique distortions, such as unrealistic objects, unnatural movements, or inconsistent visual elements, it is particularly important to accurately evaluate the quality of AIGVs. Wang et al. \cite{wang2024aigvassessorbenchmarkingevaluatingperceptual} constructed a large-scale dataset, AIGVQA-DB, which includes 36,576 AIGVs generated by 15 advanced text-to-video models using 1,048 different prompt words. Collect subjective scores from four aspects: static quality, temporal smoothness, dynamic degree, and text-video consistency, and provide pairwise comparison results between videos. 

All of the above works focus on constructing general subjective evaluation benchmarks for text-to-video generation. At the same time, there are also works that conduct subjective evaluations for specific video content types or evaluation angles. Chen et al. \cite{chen2024gaia} constructed the GAIA dataset for action quality assessment of generated videos from the perspective of actions in the generated videos, covering various full-body, hand, and facial actions. TC-Bench \cite{feng2024tc} focuses on evaluating the temporal composition and consistency of each element in the generated videos. T2VBench \cite{ji2024t2vbench} and DEVIL \cite{liao2024evaluation} evaluate from the perspective of the dynamic degree of text-to-video. T2V-CompBench \cite{sun2024t2v} evaluates T2V models from seven dimensions such as dynamic attribute binding, spatial relationship, and generation computing ability in terms of the ability of T2V models to combine different objects, attributes, and actions. In addition, Miao et al. \cite{miao2024t2vsafetybench} paid special attention to the security risks of text-to-video. They pointed out that the generated videos may contain illegal or unethical content, and previous work lacked a quantitative understanding of security. Therefore, the authors introduced T2VSafetyBench to evaluate T2V models in 12 aspects such as violence, gore, discrimination, political sensitivity, copyright, and trademark infringement. 

\subsection{Objective Metrics for T2V Generation}

Chivileva et al. \cite{chivileva2023measuring} proposed the first T2V evaluation index, which assesses the naturalness of the video and the similarity between the text and the video. In the text similarity evaluation, corresponding captions are generated for each video based on the BLIP model. Then, the similarity between the generated captions and the text prompts is calculated using the text encoder of BLIP. For the evaluation of naturalness, a series of handcrafted features are introduced, including texture, sharpness, color, spectrum, entropy, contrast, keypoint detection, and the number and size of speckles. FETV \cite{liu2024fetv} aimed at the problem that previous text-to-video evaluation metrics such as FVD, CLIPSim, and IS could not accurately reflect human subjective perception. It improved FVD and CLIPSim and proposed the index FVD-UMT for evaluating video fidelity and the index UMTScore for evaluating text-video alignment. Wu et al. \cite{wu2024towards} proposed T2VScore to automatically evaluate text-to-video in terms of video quality and text-video consistency. T2VScore consists of two components, T2VScore-A and T2VScore-Q, which evaluate text-video consistency and video quality respectively. Specifically, T2VScore-A uses LLMs to generate question-answer pairs for the prompt and then uses MLLMs to ask questions about the video and answer them. The consistency score is obtained by calculating the answer accuracy. The structure of T2VScore-Q adopts a mixture of experts structure, and the basic architecture adopts Fast-VQA. EvalCrafter \cite{liu2024evalcrafter} believes that traditional metrics such as FVD and IS cannot well fit human perception, and simple metrics cannot achieve effective evaluation. The article divides video quality, consistency, motion quality, and time consistency into fine-grained sub-dimensions, obtaining a total of 17 evaluation dimensions, and designs a dedicated evaluation algorithm for each dimension (for example, to evaluate the ability of the current T2V model to generate text, EvalCrafter directly uses the PaddleOCR toolkit \cite{ma2019paddlepaddle} to extract text from the generated video. Then, the word error rate, normalized edit distance, and character error rate are calculated and averaged). VBench \cite{huang2024vbench} also makes a finer division based on the evaluation of video quality and text-video consistency, finally obtaining 16 evaluation dimensions and corresponding 16 evaluation algorithms. Kou et al. \cite{kou2024subjective} designed the first end-to-end text-to-video evaluation model T2VQA based on the proposed T2VQA-DB dataset. It includes a video fidelity encoding module based on Siwn-3D and a text-video consistency representation module based on BLIP. Finally, the quality score is regressed through a large language model. Lu et al. \cite{lu2024aigc} improved on T2VQA by introducing a progressive training strategy, enabling different branches of the model to gradually learn relevant knowledge and achieving SOTA performance on T2VQA-DB. Li et al. \cite{zhang2024benchmarking} found that the current metrics could not comprehensively and effectively evaluate text-to-video based on the LGVQ benchmark test. Therefore, the authors proposed the UGVQ model, which systematically extracts spatial, temporal, and text-video alignment features by combining the ViT model, SlowFast network, and CLIP model, and obtains multi-dimensional scores through MLP regression. Wang et al. \cite{wang2024aigvassessorbenchmarkingevaluatingperceptual} proposed AIGV-Assessor, which utilizes spatio-temporal features and the LMM framework to capture the complex quality attributes of AIGVs, thus accurately predicting precise video quality scores and video pair preferences. 

In addition, there has been significant progress in the evaluation of Text-to-Image (T2I) generation. To address the remaining issues related to perceptual quality and text-image alignment in generated images, Wang et al. \cite{wang2025lmm4lmmbenchmarkingevaluatinglargemultimodal} proposed a comprehensive dataset and a benchmark, EvalMi-50K, for evaluating large-scale multimodal image generation. Based on EvalMi-50K, they further proposed LMM4LMM, an LMM-based metric method for evaluating large-scale multimodal T2I generation from multiple dimensions, including perception, text-image correspondence, and task-specific accuracy. Wang et al. \cite{wang2025qualityassessmentaigenerated} introduce AIGCIQA2023+, a large-scale database for AI-generated image quality assessment with multi-perspective human preference scores and detailed explanations. They propose MINT-IQA, a vision-language model that uses instruction tuning to assess and explain image quality, authenticity, and text-image correspondence, achieving state-of-the-art results on AIGC and traditional IQA benchmarks.

\subsection{Video and Image Quality Assessment}

Considerable and remarkable progress has been made in the research on traditional video and image quality assessment. Typical works such as Min et al. \cite{min2017blind} proposed a blind image quality assessment method based on pseudo-reference images (BPRI). By comparing the structural similarity between a distorted image and its extremely degraded version, BPRI achieves quality evaluation without the need for subjective score training. This method performs well across various databases and is applicable to multiple distortion types and content scenarios. Min et al. \cite{min2018blind} proposed the BMPRI method, which generates multiple pseudo-reference images by further degrading the distorted image in various ways and compares their structural similarities. Zhai and Min et al. \cite{zhai2020iqasurvey} provide a comprehensive survey of perceptual image quality assessment (IQA), reviewing major subjective IQA databases and classifying objective IQA methods, including both traditional and emerging approaches. The paper also compares state-of-the-art IQA algorithms and discusses current challenges and future trends in the field. Min et al. \cite{10.1145/3470970} present a comprehensive review of screen content quality assessment, highlighting the unique characteristics and challenges of computer-generated screen content compared to natural scenes. The paper surveys existing methodologies, benchmarks, and evaluation results, and discusses open issues and future research directions in this emerging field. Zhu et al. \cite{9961939} propose a blind image quality assessment model based on contrastive learning and cross-view consistency, integrating color and pseudo-reference information with a Transformer-based feature fusion module. The method achieves strong performance on in-the-wild IQA datasets, demonstrating improved generalization to authentic distortions. Min et al. \cite{min2024perceptualvideoqualityassessment} present an up-to-date survey of perceptual video quality assessment, summarizing key subjective and objective methods, major databases, and recent advances, including deep learning-based techniques and new application areas. Zhang et al. \cite{ZHANG2024102719} introduce UGC-New, a large-scale UGC video quality database comprising 405 original and 2430 compressed videos, with extensive subjective ratings. The authors benchmark various VQA models on this dataset and propose an improved model, New-VQA, for better quality assessment of compressed UGC videos. Min et al. \cite{min2024exploring} proposed the RichIQA method, which, for the first time, leverages rich subjective information—including mean opinion score (MOS), standard deviation of opinion scores (SOS), and distribution of opinion scores (DOS)—for image quality assessment in the wild. By adopting a three-stage network architecture and multi-label joint training, RichIQA significantly improves the modeling of subjective diversity and generalization across databases. Sun et al. \cite{sun2025empirical} proposed E-VQA, an efficient video quality assessment model based on deep learning that can strike a balance between high accuracy and low computational cost. Through a systematic analysis of the training components and extensive experiments conducted on user-generated content (UGC) datasets (KVQ, LSVQ), E-VQA demonstrated strong performance and practical efficiency, ranking third in the 2025 NTIRE UGC Video Quality Assessment Challenge. 

\section{T2VEval-Bench: A Multi-Dimensional Benchmark Dataset}

\subsection{Division of Subjective Evaluation Dimensions}

\begin{figure*}[bp]
    \centering
    \includegraphics[scale=0.8]{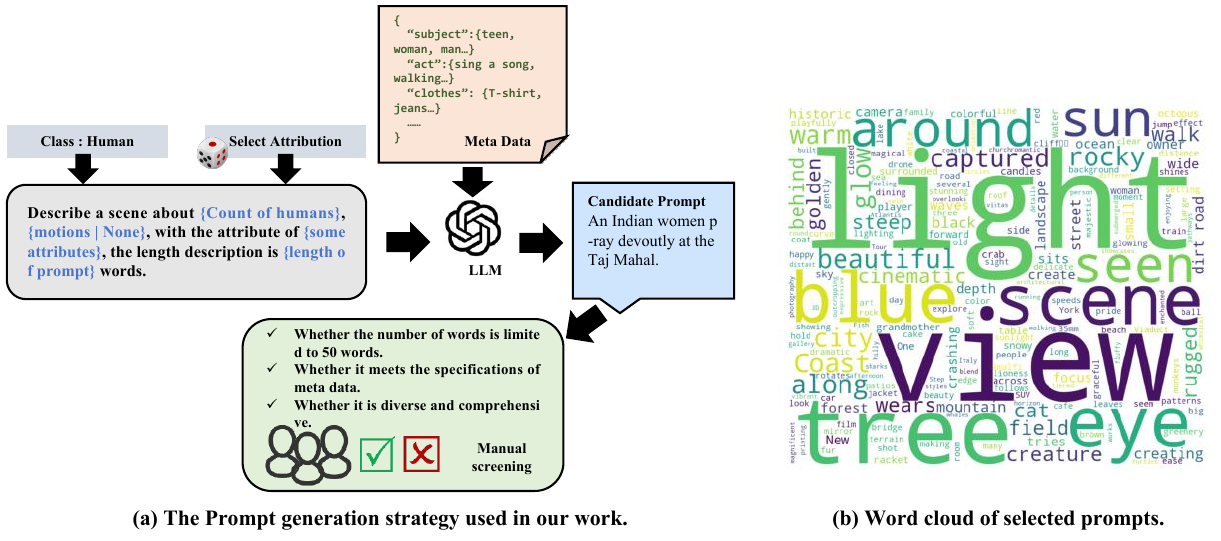}
    \caption{(a) The Prompt generation strategy used in our work. (b) The word cloud of the prompts used in T2VEval-Bench}
    \label{fig3}
\end{figure*}

\begin{figure*}[!tp]
    \centering
    \includegraphics[scale=0.8]{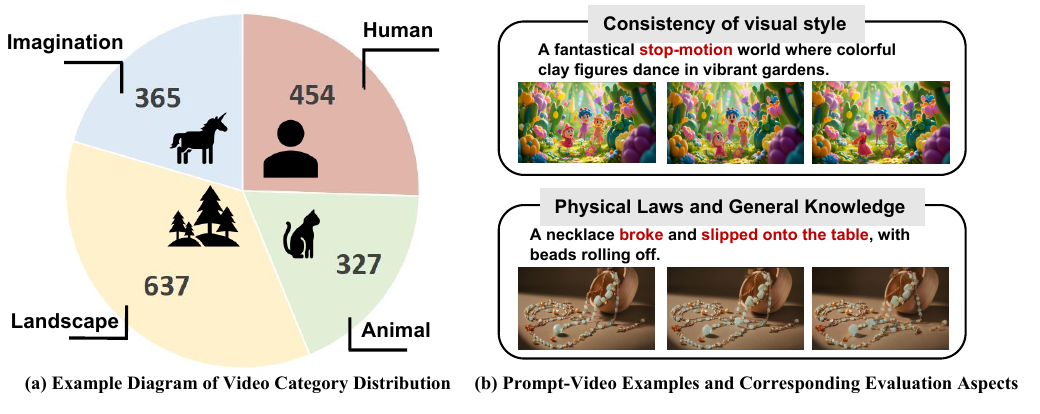}
    \caption{Prompt content category distribution and example of prompt-video pairs. (a) Category-wise distribution pie chart of T2V-generated video counts in our T2VEval-Bench. (b) Sample entries from the T2VEval-Bench containing video frames and corresponding textual prompts.}
    \label{fig4}
\end{figure*}

Traditional video quality assessment task \cite{kou2024subjective, wang2019youtube, sinno2018large, konvid1k, hosu2017konstanz} typically represents video quality using a single score, this approach may obscure the strengths and weaknesses of T2V models in specific tasks. It fails to provide a detailed analysis of the dimensions where the model performs well or poorly. Existing evaluation approaches targeting T2V models also suffer from limited applicability. To address these challenges, we proposed a multi-dimensional subjective evaluation method. Specifically, we decomposed the evaluation of text-to-video generation into four dimensions:

\begin{itemize}
\item \textbf{Overall Impression:} The evaluator’s intuitive feel and general impression of the video, serving as the video's overall score.
\item \textbf{Text-Video Consistency:} Whether the video content aligns with the given prompt.
\item \textbf{Realness:} Any content in generative videos that violates human common sense and the physical laws of the real world is not allowed. This indicator mainly quantifies whether there are situations in the video that violate physical laws (such as water not splashing when falling on the ground), physiological laws (a person having six fingers), and any human common sense (such as a sudden change in the number of subjects, inconsistent identity IDs, etc.).
\item \textbf{Technical Quality:} The performance of the video in terms of definition, dynamic smoothness, and other technical aspects.
\end{itemize}

\subsection{Compilation and Screening of Prompts}

\begin{table*}[bp]
\scriptsize
\centering
\caption{Detailed information about the models included in this article, including frame rate, duration, resolution, abilities and rease date of the version used in the experiment. *Since Sora did not open a public interface during the experiment, this paper only selects the 48 demos initially released by it for evaluation. The test of the public version of Sora will be presented in subsequent work.}
\label{tab1}
\resizebox{0.9\linewidth}{!}{      
\begin{tabular}{cccccc}\hline
\textbf{Model \& Version} & \textbf{Duration} & \textbf{FPS}& \textbf{Resolution} & \textbf{Abilities} \\
\hline
Sora* & 8-60s & 30fps & $1280\times 720$ or $1920\times 1080$ & T2V \\
Runway Gen-3 Alpha & 8s & 24fps & $2816\times 1536$ & T2V \& I2V \& V2V \\
Kling 1.0 & 5s & 30fps & $1280\times 720$ & T2V \& I2V \\
Dreamina 1.0 & 12s & 8fps & $1680\times 720$ & T2V \& I2V \\
TypeMovie & 32s & 24fps & $1024\times 576$ & T2V \& I2V  \\
Pixeling & 5s & 24fps & $1024\times 576$ & T2V \& I2V \\
Dream Machine 1.0 & 5s & 24fps & $1360\times 752$ & T2V \& I2V \\
Vega AI & 4s & 12fps & $1024\times 576$ & T2V \& I2V \\
PixVerse V2 & 4s & 18fps & $1408\times 768$ or $1024\times 576$ & T2V \& I2V\\
Pika 1.0 & 3s & 24fps & $1280\times 720$ & T2V \& I2V \\
Show-1 & 3.63s & 8fps & $576\times 320$ & T2V \\
VideoCrafter & 1.6s & 10fps & $512\times 320$ & T2V \& I2V \\
Open-Sora 1.2 & 4.25s & 24fps & $1280\times 720$ & T2V \& I2V \& V2V \\
\hline
\end{tabular}  
}                 
\end{table*}

The construction and selection of prompts are critical to the evaluation of text-to-video models, as they directly influence the credibility and reliability of the results. To ensure both diversity and comprehensiveness, we designed representative prompts aligned with evaluation dimensions and content categories. These prompts are composed of 100 self-created prompts and 48 publicly available entries from Sora’s demos.

Referring to the scheme adopted by \cite{feng2024tc, liu2024evalcrafter, huang2024vbench}, we utilized a large language model (LLM) to pre-generate prompts, followed by manual refinement, as illustrated in Figure \ref{fig3}. Basic elements for each category were predefined and organized into JSON files, which were used as inputs for the LLM to generate over 50 candidates per category, each limited to 50 words. The prompts are divided into four categories: human, animal, landscape, and imagination. We finalized 100 test prompts: \textbf{30 for human, 14 for animal, 38 for landscape, and 18 for imagination}. These prompts are designed to align with the evaluation dimension, covering aspects such as real world scenes, cultural characteristics, special weather, actions, expressions, spatial relationships and visual styles, as detailed in Figure \ref{fig4}(b).

\subsection{Video Generation}

We select 13 most representative and advanced T2V models, including Sora, Pika 1.0, Runway Gen-3 Alpha, Open-Sora 1.2, PixVerse V2, Show-1, VideoCrafter, Dreamina 1.0, Kling 1.0, Dream Machine 1.0, Pixeling, and TypeMovie. Details of these models are provided in Table \ref{tab1}. We generated a total of 1,783 videos including 48 publicly available videos from Sora and 1,735 videos generated independently. The distribution of video quantities across each category is illustrated in Figure \ref{fig4}(a).

Notably, the T2V models included in this evaluation were all released before July 2024. We plan to supplement and update the list of models in future evaluations to ensure continued relevance and comprehensiveness.

\begin{figure*}[tp]
    \centering
    \includegraphics[scale=0.45]{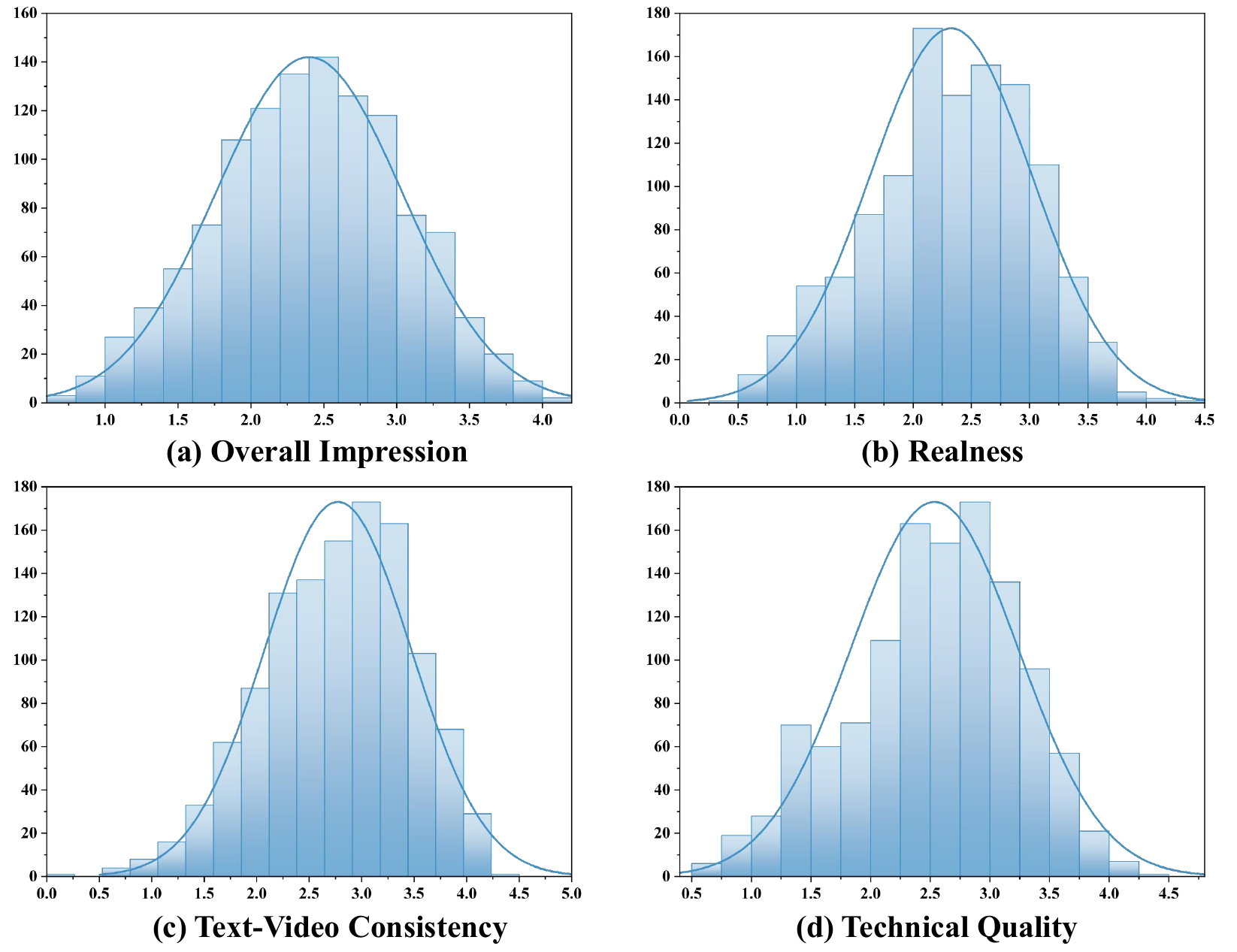}
    \caption{Distribution of subjective scores in each dimension. (a) The overall impression MOS distribution. (b) The realness MOS distribution. (c) The overall text-video consistency MOS distribution. (d) The technical quality MOS distribution.}
    \label{fig5}
\end{figure*}

\begin{figure*}[tp]
    \centering
    \includegraphics[scale=0.35]{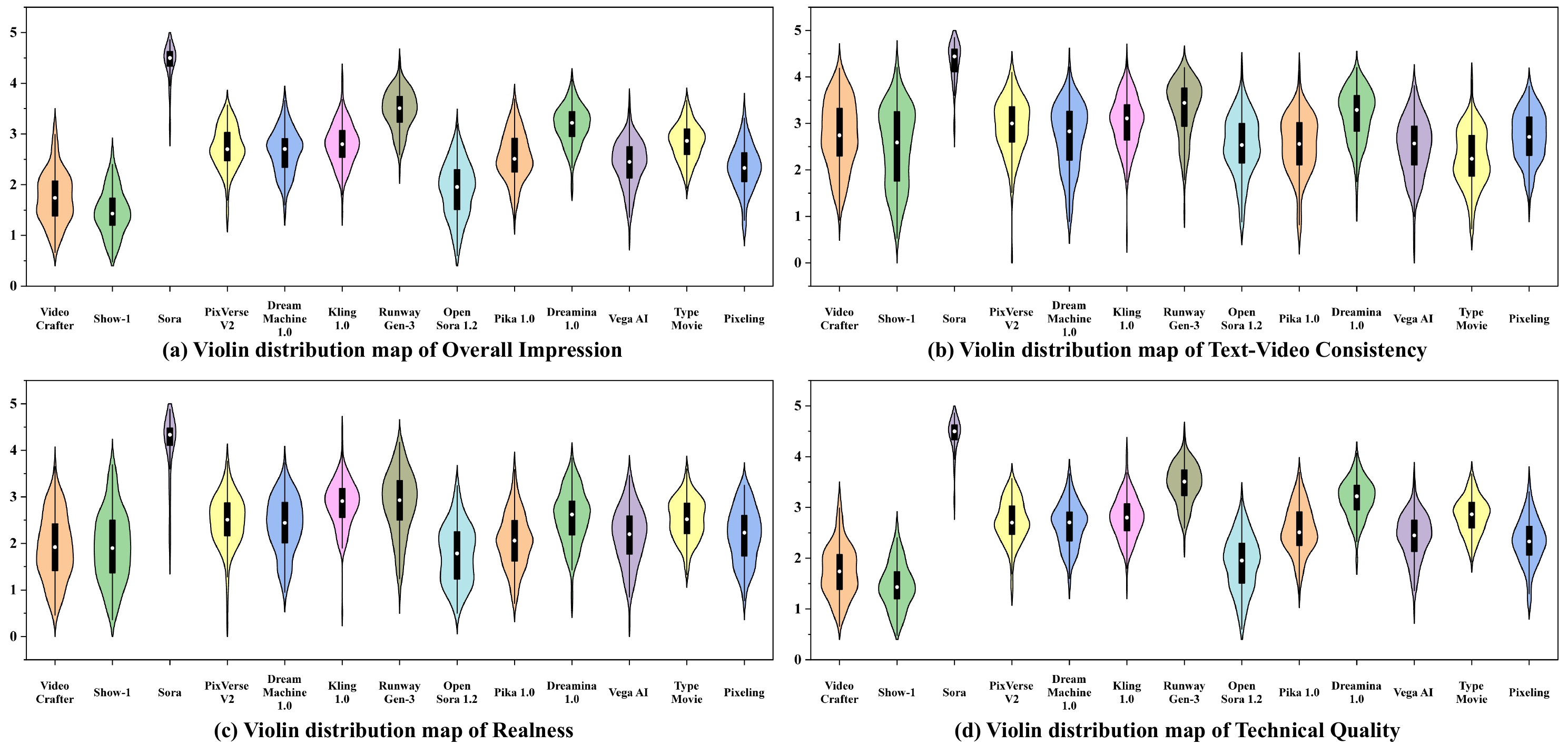}
    \caption{The Violin distribution maps of the model's subjective score in each evaluation dimension}
    \label{fig6}
\end{figure*}

\begin{figure*}[!bp]
    \centering
    \includegraphics[scale=0.45]{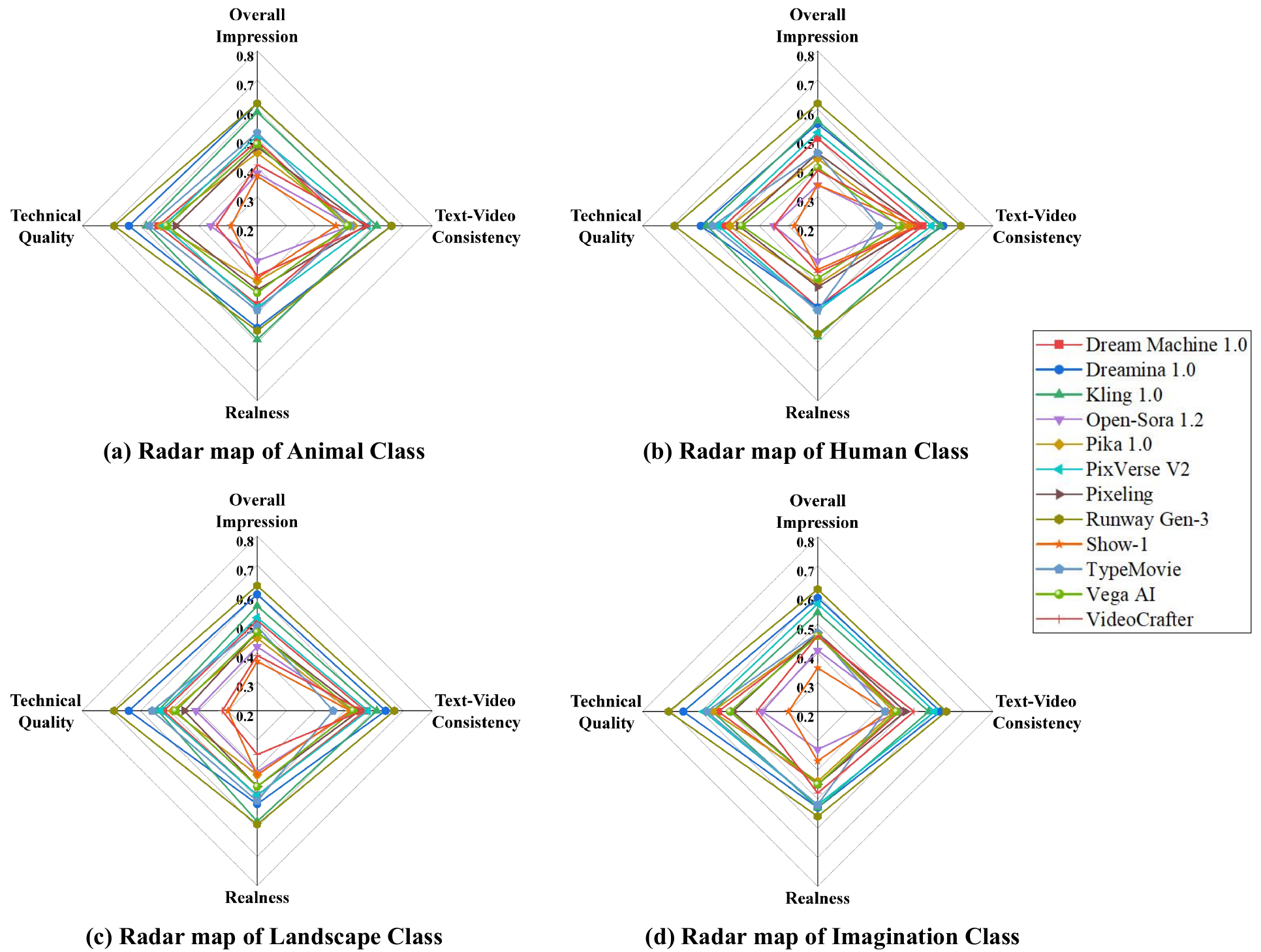}
    \caption{The radar maps of each model under different prompt's categories. Considering that Sora only had 48 demo videos, which may affect the reliability of evaluation, we did not include it in our analysis.}
    \label{fig7}
\end{figure*}

\subsection{Collection of Subjective Scores}

Subjective evaluations were conducted in accordance with the criteria of the Recommendation ITU-R BT.500 \cite{bt2019500}. The evaluation process was performed in a controlled laboratory setting to maintain consistent video viewing and evaluation conditions.  The 5-point continuous quality scale method was employed, with the scores categorized scores as follows: 0-1 (bad), 1-2 (poor), 2-3 (fair), 3-4 (good), and 4-5 (excellent). Each video was assessed by 7 to 9 professional evaluators, with a total of 73 evaluators recruited for the experiment.

We use 95\% confidence intervals to detect and remove outliers from the raw scores. For the score of a video generated by model $i$ for prompt $j$ in dimension $k$, the confidence interval is calculated as $[\mu_{ijk} - 2\sigma_{ijk}, \mu_{ijk} +2\sigma_{ijk}]$. Any scores falling outside this interval are considered outliers and removed. 

After removing outliers, we normalized the data and calculate the Mean Opinion Score (MOS) for each video. Let the video generated by model $i$ under prompt $j$ in dimension $k$ be $\textit{Video}_{ijk}$. If the score given by evaluator $l$ to $\textit{Video}_{ijk}$ is $S_{ijkl}$, then the Mean Opinion Score $\textit{MOS}_{ijk}$ of $\textit{Video}_{ijk}$ is:
\begin{equation}
    \textit{MOS}_{ijk} = \frac{1}{N} \sum_{l=1}^{N} S_{ijkl}
\end{equation}
where $N$ is the number of evaluators for $\textit{Video}_{ijk}$ after excluding outliers. Following the same method, the MOS score for each video in every dimension can be obtained. 

Figure \ref{fig5} shows the MOS distribution for each dimension, demonstrating that the scores cover nearly the entire possible range and follow a normal distribution. This confirms the comprehensiveness and validity of the subjective scores.

\definecolor{mycolor}{rgb}{0.1, 0.7, 0.1}
\begin{table*}[htbp]
\scriptsize
\centering
\caption{The rankings of the models being evaluated. The \textcolor{red}{\textbf{red}} font represents the highest score, the \textcolor{blue}{\textbf{blue}} font represents the second highest score, and the \textcolor{mycolor}{\textbf{green}} font represents the lowest score. *Sora's scores only includes 48 prompts' MOSs, which may have certain limitations.}
\label{tab2}
\resizebox{1.0\linewidth}{!}{      
\begin{tabular}{ccccc}\hline
\textbf{Model} & \textbf{Overall Impression} & \textbf{Text-Video Consistency}& \textbf{Realness} & \textbf{Technical Quality} \\
\hline
Sora* & \textcolor{red}{\textbf{0.851}} &	\textcolor{red}{\textbf{0.864}} &	\textcolor{red}{\textbf{0.823}} &	\textcolor{red}{\textbf{0.882}} \\
Runway Gen-3 & \textcolor{blue}{\textbf{0.633}} &	\textcolor{blue}{\textbf{0.668}} &	\textcolor{blue}{\textbf{0.574}} &	\textcolor{blue}{\textbf{0.695}}   \\
Dreamina 1.0 & 0.590 &	0.629 &	0.512 &	0.633   \\
Kling 1.0 & 0.565 &	0.608 &	0.572 &	0.557   \\
Pixverse V2 & 0.531 &	0.589 &	0.498 &	0.539\\
Dream Machine 1.0 & 0.496 &	0.543 &	0.478 & 0.528  \\
TypeMovie & 0.483 &	\textcolor{mycolor}{\textbf{0.451}} & 0.500 & 0.560  \\
Pixeling & 0.466 &	0.544 &	0.420 &	0.464 	  \\
Vega AI & 0.453 &	0.496 &	0.427 &	0.481  \\
Pika 1.0 & 0.445 &	0.499 &	0.407 &	0.504	\\
VideoCrafter & 0.410 &	0.545 &	0.381 &	0.347   \\
Open-Sora 1.2 & 0.392 &	0.500 &	\textcolor{mycolor}{\textbf{0.350}} &	0.375 \\
Show-1 & \textcolor{mycolor}{\textbf{0.356}} & 0.486 & 0.385 &	\textcolor{mycolor}{\textbf{0.286}}   \\
\hline
\end{tabular}  
}                 
\end{table*}

\subsection{Analysis of T2V Model Performance}

To analyze the performance of T2V models, we calculated the MOS scores of each model in every dimension. Let the mean opinion score of model \( i \) in dimension \( k \) be \(\textit{MOS}_{ik}\), then the \(\textit{MOS}_{ik}\) can be calculated as follows:
\begin{equation}
\textit{MOS}_{ik} = \frac{1}{M} \sum_{j=1}^{M} \textit{MOS}_{ijk}
\end{equation}
where \( M \) is the number of prompts. Table \ref{tab2} provides the MOS scores of the 13 models across all evaluation dimensions, while Figure \ref{fig6} visualizes the score distribution using a violin plot. Notably, Sora achieves the highest scores in all dimensions, exhibiting a highly concentrated and stable distribution. However, it is important to note that Sora’s ranking is based solely on its 48 demo videos, which introduces certain limitations to the evaluation.

Aside from Sora, Ruway Gen-3 demonstrates strong competitiveness and potential across all dimensions and Kling 1.0 excels in the realness dimension. In contrast, Open-Sora 1.2 and Show-1 exhibit the weakest performance, primarily due to their poor technical quality with noticeable artifacts like block distortions. These models also struggle with maintaining entity consistency (e.g., people and animals) and temporal coherence.

\begin{figure}[!h]
    \centering
    \includegraphics[scale=0.55]{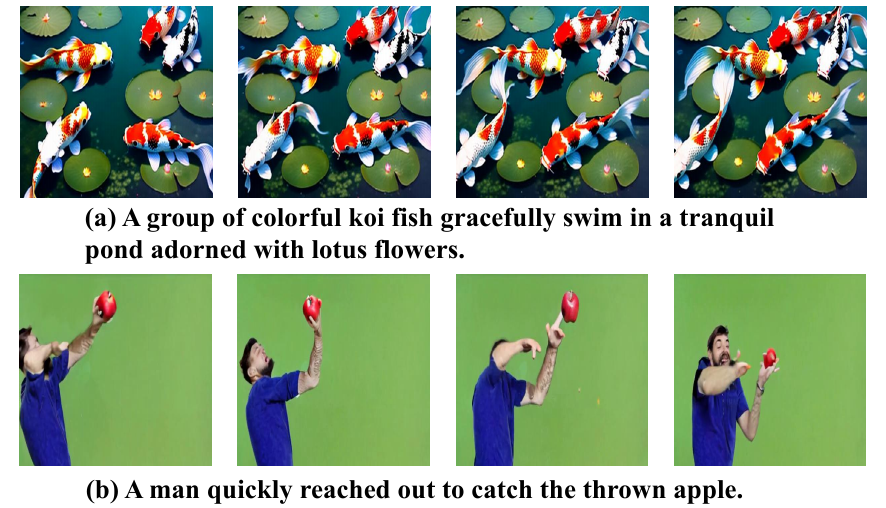}
    \caption{Examples of realness distortion in T2V-generated videos. (a) It is impossible to simulate the natural feeling of fish swimming in water. (b) The limbs of the person are distorted and the action of "catching an apple" is not expressed.}
    \label{fig8}
\end{figure}

Finally, we analyzed the performance of these models in each category, and the results are shown in  Figure \ref{fig7}. Runway Gen-3 consistently outperforms other models across all prompt categories, while Open-Sora 1.2, VideoCrafter, and Show-1 exhibit the weakest performance. This aligns with the conclusions presented in Table \ref{tab2}. Furthermore, it is evident that regardless of the prompt category, the scores for the realness dimension are generally lower than those of other dimensions. This highlights that the primary challenge faced by current T2V models lies in accurately understanding and representing the objective laws of the real world, including physical principles and cognitive commonsense knowledge. Specifically, these models struggle to accurately understand and represent the physiological structures (e.g., facial features, limbs), movement patterns (e.g., running, dancing), and physical laws (e.g., sliding, dripping) of organisms in the real world, as illustrated in Figure \ref{fig8}.

\section{T2VEval: a Multi-Branch Fusion Scheme for Objective Evaluation}

\begin{figure*}[tp]
    \centering
    \includegraphics[scale=0.5]{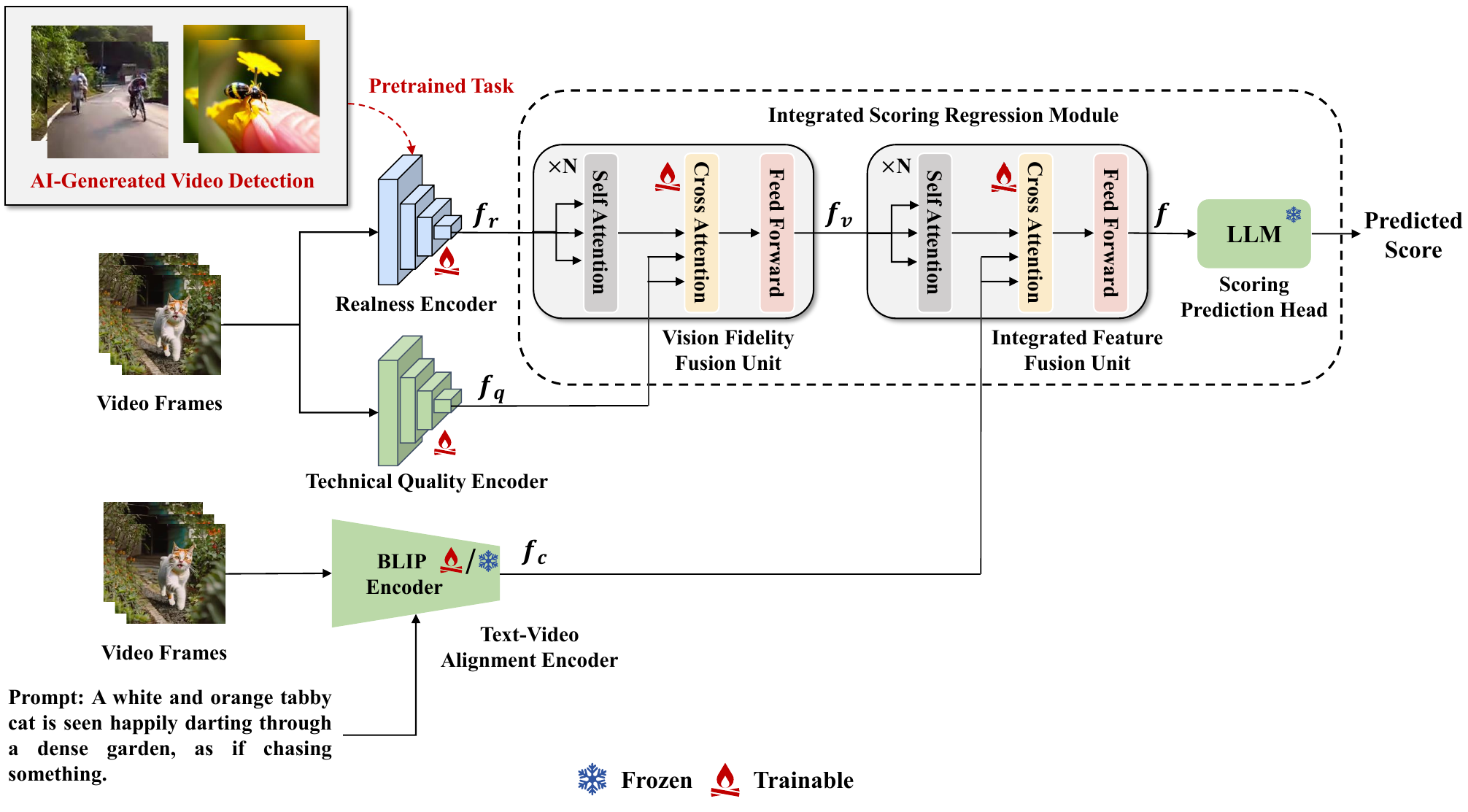}
    \caption{The overall architecture of proposed T2VEval.}
    \label{fig9}
\end{figure*}

Based on the T2VEval-Bench, we proposed a multi-branch fusion framework for objective evaluation called \textbf{T2VEval}, which employs a multi-task architecture to accomplish the quality evaluation of T2V-generated videos. The framework comprises three parallel encoding branches—a technical quality encoder, a realness encoder, and a text-video alignment encoder—coupled with an integrated score prediction module. This prediction module incorporates three key components: a vision fidelity fusion unit, an integrated feature fusion unit, and a scoring prediction head. Figure \ref{fig9} illustrates the architecture of T2VEval. 

\subsection{Technical Quality Encoder}

The technical quality encoder aims to represent distortions in both the spatial and temporal domains, including noise, overexposure/underexposure, motion blur, jitter etc. This part can be understood as a conventional video quality assessment (VQA) problem. Recent studies have demonstrated that the Swin-T 3D architecture is well-suited for a wide range of VQA tasks \cite{wu2022fast, wu2023exploring}. Leveraging its 3D sliding window attention mechanism, the model excels at extracting spatio-temporal features, enabling effective analysis of video distortions across both spatial and temporal domains. Therefore, we directly adopt the Tec branch from DOVER \cite{wu2023exploring} as the backbone of our technical quality encoder. Given a video clip $v$, we can extract its corresponding technical quality feature $f_q$: 
\begin{equation}
f_q = Swin3D(v)
\end{equation}

\subsection{Realness Encoder}

Current videos generated by T2V modols commonly suffer from realness issues, including missing the limbs of persons or animals, distorted facial structures, inconsistent inter-frame content, and unnatural motion patterns. These flaws essentially stem from the generated content contradicting cognitive expectations, which are inherently complex and challenging to model using simplistic network architectures.

Inspired by natural language processing work such as BERT \cite{devlin2018bert} and PPT \cite{gu2021ppt}, we designed a pre-training task for the realness evaluation, named AI-generated videos detection. We constructed a pre-training dataset comprising 5,000 AI-generated videos and 5,000 human-made videos, sourced from the DeMemba datasets \cite{chen2024demamba}. Using this dataset, we trained a classification network with ConvNeXt 3D \cite{liu2022convnet} as the backbone architecture. This network achieves a classification accuracy of 98\% on the training dataset and an 83\% zero-shot classification accuracy on the T2VEval-Bench dataset. We believe that the AI-generated video detection task and the T2V-generated video realness evaluation task exhibit stronger correlations. Compared to conventional ImageNet-based pre-training method, our pre-training strategy enables the model to better understand and characterize the realness flaws in T2V-generated videos. For video clip $v$, the realness encoder can extract the corresponding realness feature $f_r$: 
\begin{equation}
f_r=ConvNeXt3D(v)
\end{equation}

\subsection{Text-Video Alignment Encoder}

Text-video alignment refers to the consistency between the video content and the text prompt. CLIP \cite{radford2021learning} and BLIP \cite{li2022blip} are models with strong zero-shot text-to-image matching capabilities. However, experiments \cite{chivileva2023measuring, huang2024vbench, liu2024evalcrafter, liu2024fetv} have shown that simply using CLIP or BLIP has a low correlation with the subjective scores. Additionally, some approaches attempt to generate captions for the video using MLLM (Multimodal Large Language Models) and then compare the similarity between the captions and prompts \cite{chivileva2023measuring}, but results indicate this method is also unreliable.

Therefore, we considered modeling text-video alignment in the latent space. Specifically, we used a frozen BLIP encoder to separately encode the text and video frames, and then project the text and video features into a unified feature space through a feature alignment layer. Finally, a cross-attention module was used to encode the similarity between the text and video features. Given a video $v$ and text prompt $t$, the encoder extracts the consistency feature $f_c$:
\begin{equation}
f_c= SA(CA\{BLIP(v_i),BLIP(t)\}_{i=1}^N)
\end{equation}

\subsection{Integrated Scoring prediction Module}

The proposed \textbf{Integrated Scoring Regression Module} consists of three key components: \textbf{the vision fidelity fusion unit, the integrated feature fusion unit, and the scoring prediction head}. Inspired by T2VQA \cite{wu2024towards}, the module adopts a Transformer-based architecture to achieve a progressive and effective integration of features.

First, the vision fidelity fusion unit focuses on fusing the technical quality feature $f_q$ and the realness feature $f_r$, resulting in the vision fidelity feature $f_v$ that capture the visual defects of the videos. This step leverages the strong task dependency between video realness and technical quality, both of which are entirely reliant on visual content. 
\begin{equation}
f_v=FFN(SA(CA(f_r), f_q))
\end{equation}

Then the integrated feature fusion unit, furthermore, fuses $f_v$ with $f_c$ to obtain the integrated feature from the visual-textual perspective.

\begin{equation}
f = FFN(SA(CA(f_v), f_c))
\end{equation}

LLM has achieved excellent results on IQA and VQA tasks \cite{wu2024q,wu2023q,wu2025towards,zhang2024q}. Inspired by this, we employed LLM as the scoring prediction head in T2VEval and guided LLM by designing a special prompt "Please rate the quality of the video". The output of LLM is the probability distribution of the five quality levels (bad, poor, fair, good, and excellent). We assigned weights of 1-5 to these five quality levels. Finally, the predicted score was obtained by calculating the softmax of each token and multiplying it by the corresponding weight.
\begin{equation}
s_{pred}=\sum_{i=1}^5i\times softmax(\lambda_i)=\sum_{i=1}^5i\times\frac{e^{\lambda_i}}{\sum_{j=1}^5e^{\lambda_j}}
\end{equation}
where $\lambda_i$ is the probability distribution of the i-th level token.

\begin{table*}[bp]
\centering
\caption{Performance comparison of different models. \textbf{\textcolor{red}{Red}}: the best}
\scalebox{0.95}{%
\begin{tabular}{ccccccccccc}
\toprule
\multirow{2}{*}{\textbf{Type}} & \multirow{2}{*}{\textbf{Models}} & \multicolumn{4}{c}{\textbf{T2VEval-Bench}} & \multicolumn{4}{c}{\textbf{T2VQA-DB}} \\
\cmidrule(lr){3 - 6} \cmidrule(lr){7 - 10}
& & \textbf{SROCC $\uparrow$} & \textbf{PLCC $\uparrow$} & \textbf{KROCC $\uparrow$} & \textbf{RMSE $\downarrow$} & \textbf{SROCC $\uparrow$} & \textbf{PLCC $\uparrow$} & \textbf{KROCC $\uparrow$} & \textbf{RMSE $\downarrow$} \\
\midrule
\multirow{4}{*}{\textbf{zero-shot}} & CLIPSim & 0.0460 & 0.0656 & 0.0296 & 0.9171 & 0.1047 & 0.1277 & 0.0702 & 21.683 \\
& BLIP & 0.1690 & 0.1718 & 0.1116 & 0.9131 & 0.1659 & 0.1860 & 0.1112 & 18.373 \\
& ImageReward & 0.2611 & 0.2798 & 0.1738 & 0.8515 & 0.1875 & 0.2121 & 0.1266 & 18.243 \\
& UMTScore & 0.1832 & 0.1600 & 0.1258 & 0.9196 & 0.0676 & 0.0721 & 0.0453 & 22.559\\
\midrule
\multirow{5}{*}{\textbf{fine-tuned}} & SimpleVQA & 0.4815 & 0.5206 & 0.3421 & 0.6031 & 0.6275 & 0.6338 & 0.4466 & 11.163 \\
& FAST-VQA & 0.5878 & 0.5952 & 0.4382 & 0.6660 & 0.7173 & 0.7295 & 0.5303 & 10.595 \\
& DOVER & 0.6070 & 0.5994 & 0.4272 & 0.6107 & 0.7609 & 0.7693 & 0.5704 & 9.8072\\
& T2VQA & 0.6598 & 0.6603 & 0.4757 & 0.5749 & 0.7965 & 0.8066 & 0.6058 & 9.0210 \\
&\textbf{\textcolor{red}{T2VEval (Ours)}}& \textbf{\textcolor{red}{0.7098}} & \textbf{\textcolor{red}{0.7252}} & \textbf{\textcolor{red}{0.5218}} & \textbf{\textcolor{red}{0.5296}} & \textbf{\textcolor{red}{0.8049}} & \textbf{\textcolor{red}{0.8175}} & \textbf{\textcolor{red}{0.6159}} & \textbf{\textcolor{red}{8.6133}} \\
\bottomrule
\end{tabular}
}
\label{tab4}
\end{table*}

\subsection{Training Strategy}

To maximize the effectiveness of each branch and enable them to focus on task-specific learning, we proposed a divide-and-conquer training strategy. This approach ensures each branch specializes in its designated task while contributing to the overall performance.

Initially, we prioritized enhancing features unrelated to text, such as technical quality and realness, by refining purely visual dimensions without involving text input. In the next phase, we fine-tuned the full parameters of the BLIP model to optimize text-video alignment. Finally, we integrated the three branches with a fusion unit, fine-tuning a limited set of parameters to promote cooperative learning across multiple aspects, including technical quality, realness, and text-video alignment.

Specifically, we first fine-tuned all parameters of the technical quality branch using the technical quality MOS scores, and similarly fine-tuned all parameters of the realness branch using the realnes MOS scores. For the alignment branch, the MOS scores of the text-video consistency were used to fine-tune all parameters of the BLIP model. In the final stage, we incorporated the fusion units and selectively fine-tuned a subset of the BLIP parameters.

To further enhance performance, we adopted a dual-loss optimization strategy \cite{li2022blindly} for each branch. This strategy combines Pearson's Linear Correlation Coefficient (PLCC) and ranking loss, ensuring robust optimization tailored to each evaluation dimension.

\begin{equation}
L_{plcc} = \frac{1}{2} \left( 1 - \frac{\sum_{i=1}^m (s_i - \bar{s})(y_i - \bar{y})}{\sqrt{\sum_{i=1}^m (s_i - \bar{s})^2} \sqrt{\sum_{i=1}^m (y_i - \bar{y})^2}} \right)
\end{equation}

\begin{equation}
L_{rank} = \frac{1}{m^2} \sum_{i=1}^m \sum_{j=1}^m \left( \max\left( 0, |y_i - y_j| - e(y_i, y_j) \cdot (s_i - s_j) \right) \right)
\end{equation}

Here, $y$ and $\bar{y}$ denote the MOSs and their mean value, respectively, while $s$ and $\bar{s}$ represent the prediction scores and corresponding mean value. 

The final loss function is defined as:
\begin{equation}
L = L_{plcc} + \lambda \cdot L_{rank}
\end{equation}
$\lambda$ is a hyperparameter used to balance the different loss functions, which we set to 0.3.

\section{Experiments and results}

\subsection{Implement Details}

When training and testing on T2VEval-Bench, we followed the common practice of dataset splitting by leaving out 80\% for training, and 20\% for testing. To eliminate bias from a single split, we randomly split the dataset 10 times and averaged the results for performance comparison.

For the text-video alignment branch, we used the BLIP encoder \cite{li2022blip} to extract features, while the technical quality encoder was initialized with weights pre-trained by Swin-T on the Kinetics-400 dataset \cite{kay2017kinetics}. For the realness encoder, we initialized it with pre-trained parameters on a self-built classification dataset. The LLM used in the scoring prediction head is the 7B version of Vicuna v1.5 \cite{zheng2023judging}, which is fine-tuned from Llama2 \cite{touvron2023llama}. What needs illustration is that the gradients of the Vicuna v1.5 were frozen during training.

In the training and testing process, we first sampled 8 frames from the input video and resized them to $224 \times 224$. The Adam optimizer was used, which was initialized with a learning rate of 1e-5, and the learning rate decayed from 1 to 0 using a cosine scheduler. T2VEval was trained for 30 epochs with a batch size of 4 on a server equipped with one Nvidia GeForce RTX 3090.

The evaluation metrics include the Pearson Linear Correlation Coefficient (PLCC), Spearman Rank-Order Correlation Coefficient (SROCC), Kendall’s Rank-Order Correlation Coefficient (KROCC) and Root Mean Square Error (RMSE). SROCC and PLCC measure the monotonicity of the predictions, while RMSE assesses prediction accuracy. A better model should exhibit higher SROCC, PLCC and KROCC values, whereas lower RMSE values indicate better performance.

\begin{figure*}[bp]
    \centering
    \includegraphics[scale=0.22]{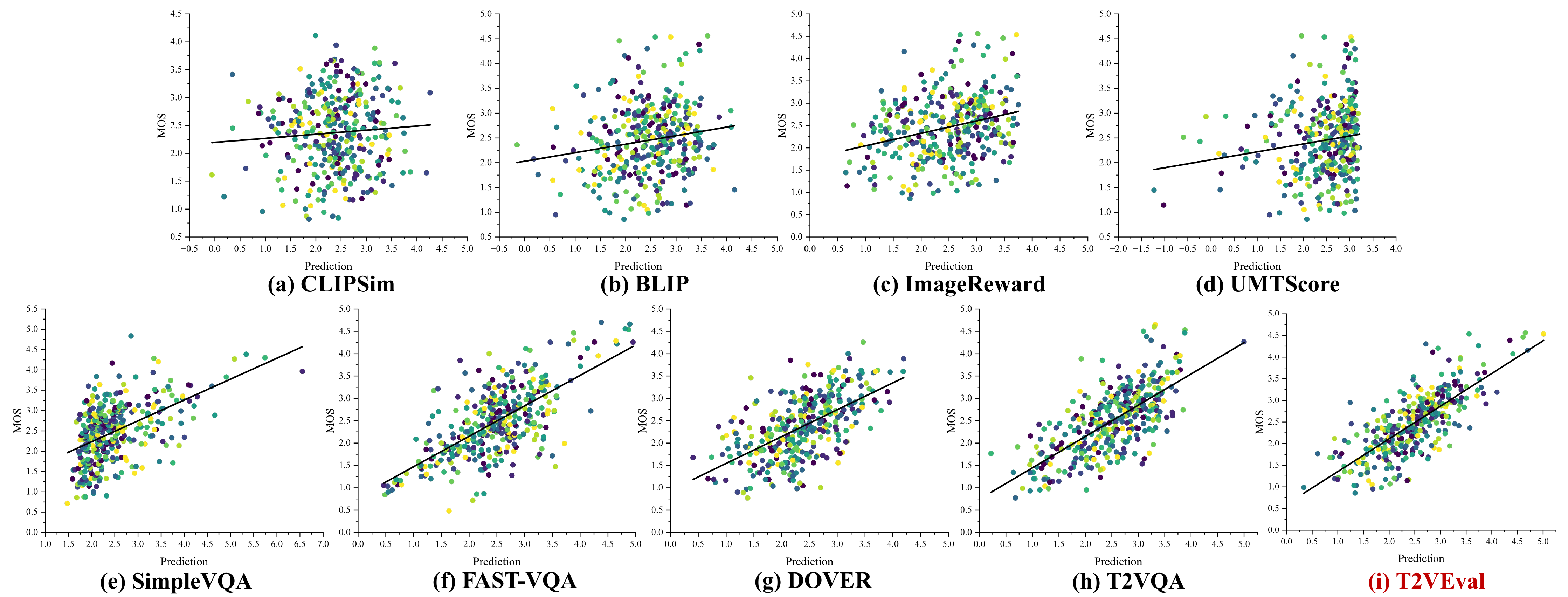}
    \caption{Scatter plots of the predicted scores vs. MOSs. The curves are obtained by a four-order polynomial nonlinear fitting. The brightness of scatter points from dark to bright means density from low to high.}
    \label{fig10}
\end{figure*}

\subsection{Performance Comparison}

We utilized CLIPSim \cite{wu2021godiva}, BLIP \cite{li2022blip}, ImageReward \cite{xu2024imagereward}, UMTScore \cite{liu2024fetv}, Simple VQA \cite{sun2022deep}, Fast-VQA \cite{wu2022fast}, DOVER \cite{wu2023exploring}, and T2VQA \cite{wu2024towards} as reference algorithms for performance comparison experiments. CLIPSim computes the average similarity between text and individual video frames. BLIP and ImageReward are adapted to convert image quality assessment (IQA) metrics into VQA metrics, while UMTScore is tailored to evaluate text-video alignment. T2VQA is specifically designed for T2V evaluation. DOVER, Simple VQA, and Fast-VQA are effective models for VQA. We employed pre-trained weights for CLIPSim, BLIP, ImageReward, and UMTScore for zero-shot testing, while Simple VQA, Fast-VQA, DOVER, and T2VQA are fine-tuned on the T2VEval-Bench.

Table \ref{tab4} compares the performance of T2VEval with other methods. The results demonstrate that our method achieves state-of-the-art performance on T2VEval-Bench, surpassing the second-best method by nearly 4\%. The performance of zero-shot models is notably poor, which may be attributed to their sole focus on text-video alignment and inability to capture temporal information between video frames. Furthermore, these methods are typically trained and tested on datasets with human-made videos, overlooking the domain gap between AI-generated videos and human-made videos. Although VQA models achieve higher scores, they still fall significantly short of T2VQA and T2VEval, further underscoring the necessity of multi-branch fusion evaluation framework evaluation frameworks.

Figure \ref{fig10} shows scatter plots comparing predicted scores and MOS scores for BLIP, CLIPSim, ImageReward, UMTScore, Simple VQA, Fast-VQA, DOVER, T2VQA, and T2VEval. A better-performing model exhibits a fitting curve closer to the diagonal with fewer scattered points. As depicted, T2VEval significantly outperforms other algorithms, demonstrating a closer fit and reduced dispersion in predictions.

In addition, to verify the effectiveness and advancement of T2VEval, we fine-tuned and tested T2VEval on the T2VQA-DB, LGVQ, and Q-Eval datasets. Table \ref{tab4} shows the experimental results on T2VQA-DB, where T2VEval also achieved state-of-the-art performance on this dataset. Table \ref{LGVQ} shows the performance of T2VEval on LGVQ. It can be seen that T2VEval achieved the best performance in both the temporal quality and consistency dimensions, but was slightly insufficient in spatial quality. This may be because our backbone networks preferentially adopt 3D spatio-temporal modeling networks, which also takes into account the biggest difference between videos and images: videos are dynamic and change over time. Table \ref{tab_Q_Eval} shows the performance of our method on Q-Eval, and our method also achieved state-of-the-art performance. It should be noted that since the complete test set of Q-Eval cannot be obtained, we can only perform dataset partitioning and testing on the publicly available training set.

\begin{table}[ht]
\centering
\caption{Performance comparison on LGVQ. \textbf{\textcolor{red}{Red}}: the best}
\begin{tabular}{llccc}
\toprule
\textbf{Aspects} & \textbf{Methods} & \textbf{SROCC} & \textbf{PLCC} & \textbf{KROCC} \\
\midrule
\multirow{7}{*}{Spatial}
    & MUSIQ           & 0.669 & 0.682 & 0.491 \\
    & StairIQA        & 0.701 & 0.737 & 0.521 \\
    & CLIP-IQA        & 0.684 & 0.709 & 0.502 \\
    & LIQE            & 0.721 & 0.752 & 0.538 \\
    & UGVQ            & 0.759 & 0.795 & 0.567 \\
    & AIGV-Assessor   & \textbf{\textcolor{red}{0.803 }}& \textbf{\textcolor{red}{0.819}} & \textbf{\textcolor{red}{0.617}} \\
    & T2VEval         & 0.753 & 0.789 & 0.560 \\
\midrule
\multirow{5}{*}{Temporal}
    & VSFA            & 0.841 & 0.857 & 0.643 \\
    & SimpleVQA       & 0.857 & 0.867 & 0.659 \\
    & FastVQA         & 0.849 & 0.843 & 0.647 \\
    & DOVER           & 0.867 & 0.878 & 0.672 \\
    & UGVQ            & 0.893 & 0.907 & 0.703 \\
    & AIGV-Assessor   & 0.900 & 0.902 & 0.717 \\
    & \textbf{T2VEval}         & \textbf{\textcolor{red}{0.906}} & \textbf{\textcolor{red}{0.914}} & \textbf{\textcolor{red}{0.720}} \\
\midrule
\multirow{8}{*}{Alignment}
    & CLIPScore       & 0.446 & 0.453 & 0.301 \\
    & BLIPScore       & 0.455 & 0.464 & 0.319 \\
    & ImageReward     & 0.498 & 0.499 & 0.344 \\
    & PickScore       & 0.501 & 0.515 & 0.353 \\
    & HPSv2           & 0.504 & 0.511 & 0.357 \\
    & UGVQ            & 0.551 & 0.555 & 0.394 \\
    & AIGV-Assessor   & 0.577 & 0.578 & 0.411 \\
    & \textbf{T2VEval}         & \textbf{\textcolor{red}{0.674}} & \textbf{\textcolor{red}{0.694}} & \textbf{\textcolor{red}{0.496}} \\
\bottomrule
\end{tabular}
\label{LGVQ}
\end{table}

\begin{table}[ht]
\centering
\caption{Performance comparison on Q-Eval. \textbf{\textcolor{red}{Red}}: the best}
\begin{tabular}{llcc}
\toprule
\textbf{Aspects} & \textbf{Methods} & \textbf{PLCC} & \textbf{SROCC} \\
\midrule
\multirow{5}{*}{Visual Quality}
    & SimpleVQA   & 0.346  & 0.385  \\
    & DOVER           & 0.506 & 0.506 \\
    & Q-Align        & 0.416  & 0.460  \\
    & T2VQA        & 0.516  & 0.516  \\
    & \textbf{T2VEval}        & \textcolor{red}{\textbf{0.535}}  & \textcolor{red}{\textbf{0.517}}  \\
\midrule
\multirow{4}{*}{Alignment}
    & CLIPScore      & 0.443  & 0.431  \\
    & BLIP2Score     & 0.488  & 0.483  \\
    & ImageReward    & 0.485  & 0.472  \\
    & \textbf{T2VEval}        & \textcolor{red}{\textbf{0.685}}  & \textcolor{red}{\textbf{0.688}}  \\
\bottomrule
\end{tabular}
\label{tab_Q_Eval}
\end{table}

\subsection{Ablation Studies}

To systematically evaluate the contribution of core architectural components in T2VEval, we conducted comprehensive ablation studies. Specifically, in Table \ref{tab5}, we conducted experiments to evaluate (i) different vision fidelity fusion mechanisms and (ii) differential pre-training strategies applied to the realness encoder. Additionally, in Table \ref{tab6}, we explored the contribution and effect of each branch, aiming to gain a deeper understanding of how these components interact and influence the overall performance of T2VEval.

\subsubsection{\textbf{Ablation Experiment of Vision Fidelity Fusion Strategy}}

In T2VEval, an effective feature fusion strategy is required to integrate features from multiple branches. The design of this strategy involves two critical aspects: the selection of fusion mechanisms and the order of combining features derived from different branches. 

T2VEval employs a cross-attention mechanism to fuse technical quality and realness features. In previous studies, alternative fusion methods, such as feature summation and concatenation, have also been utilized. To evaluate their effectiveness, we compared the cross-attention fusion strategy against addition and concatenation methods (Addition is denoted as \textbf{T2VEval-Add}, and concatenation is denoted as \textbf{T2VEval-Cat}). As shown in Table \ref{tab5}, the cross-attention method significantly outperforms the other two approaches.

\begin{figure*}[tbhp]
    \centering
    \includegraphics[scale=0.35]{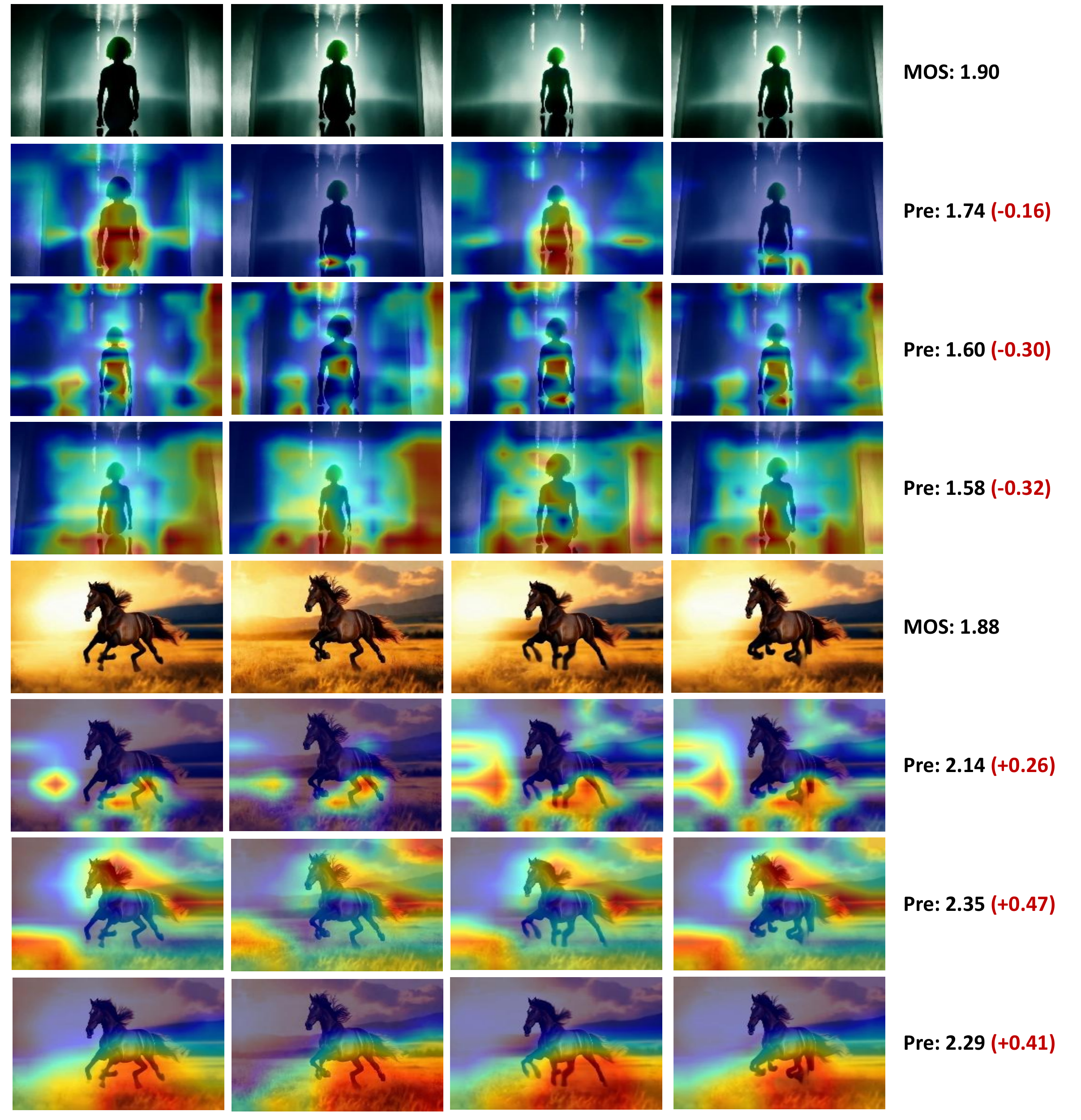}
    \caption{Grad-CAM diagram of 3D network under different pre-training strategies and the final T2VEval.}
    \label{fig11}
\end{figure*}

This superiority can be explained by the dynamic nature of cross-attention, which assigns adaptive weights to different features based on their intrinsic relevance. Unlike summation and concatenation, cross-attention selectively highlights valuable information while maintaining contextual integrity. Feature summation, on the other hand, may result in information loss due to the "smoothing" effect during addition, which reduces the distinctions between features. While concatenation preserves all feature information, it fails to establish meaningful interactions between features.

In contrast, cross-attention enables pairwise interactions and weight calculations for features, effectively preserving and enhancing information integrity. This mechanism allows the model to capture more complex relationships between features, ensuring that key information is neither diluted nor masked, ultimately leading to superior performance.

A well-designed and accurate order of feature fusion is crucial for capturing the complementary information between features from different branches while maintaining their independence. To explore the optimal fusion order, we compared several different fusion strategies, including the one used in our T2VEval scheme:

\begin{itemize}
    \item First, fuse the technical quality feature $f_q$ and the consistency feature $f_c$. Then, the fusion result is fused with the realness feature $f_r$. This process is denoted as \textbf{T2VEval-QC\&R}.
    \item First, fuse the realness feature $f_r$ and the consistency feature $f_c$. Then, the fusion result is fused with the technical quality feature $f_q$. This process is denoted as \textbf{T2VEval-RC\&Q}.
    \item First, fuse the realness feature $f_r$ and the technical quality feature $f_q$ with the consistency feature $f_c$ separately. Then, fuse the two fusion results. This process is denoted as \textbf{T2VEval-RC\&QC}.
    \item First, fuse the realness feature $f_r$ and the technical quality feature $f_q$. Then the fusion result is fused with the consistency feature $f_c$. This process is denoted as \textbf{T2VEval-RQ\&C} and is adopted in our T2VEval scheme.
\end{itemize}

Results are shown in Table \ref{tab5}. These results may be due to the fact that realness and technical quality are completely dependent on visual content and have a stronger task-dependence. In contrast, due to the existence of the text modality, fusing visual-related features (whether realness feature or technical quality feature) with consistency features first may lead to interference among various modal features.

\begin{table*}[htbp]
\caption{Performance comparison of ablation studies. \textbf{\textcolor{red}{Red}}: the best}
\centering
\scalebox{1.0}{
\begin{tabular}{cccccc}
    \toprule
    \textbf{Categories of ablation experiments} & \textbf{Models} & \textbf{SROCC $\uparrow$} & \textbf{PLCC $\uparrow$} & \textbf{KROCC $\uparrow$} & \textbf{RMSE $\downarrow$} \\
    \midrule
    \multirow{5}{*}{Vision Fidelity Fusion Strategy} 
    & T2VEval-QC\&R & 0.6826  & 0.6946 & 0.5001 & 0.5826\\
    & T2VEval-RC\&Q & 0.6838 & 0.6691 & 0.4925 & 0.5664 \\
    & T2VEval-RC\&QC & 0.6757 & 0.6680 & 0.4833 & 0.5988 \\
    & T2VEval-Cat & 0.6524 & 0.6493 & 0.4680 & 0.5857 \\
    & T2VEval-Add & 0.6678 & 0.6609 & 0.4867 & 0.5782\\
    \midrule
    \multirow{2}{*}{Pre-train Strategy for Realness Branch} 
    & T2VEval-Random & 0.6683 & 0.6443 & 0.4630 & 0.5721 \\
    & T2VEval-ImageNet & 0.6704 & 0.6760 & 0.4901 & 0.5883 \\
    \midrule
    \multirow{1}{*}{Effectiveness of Each Branch and Training Strategy} 
    & T2VEval-Once & 0.6977 & 0.6860 & 0.5001 & 0.5553 \\
    \midrule
    \multirow{2}{*}{Scoring Prediction Head} 
    & T2VEval-Linear & 0.6754 & 0.6919 & 0.4774 & 0.5512 \\
    & T2VEval-Conv1d & 0.6978 & 0.7089 & 0.4945 & 0.5483 \\
    \midrule
    \multirow{1}{*}{\textcolor{red}{\textbf{Our Method}}}
    & \textcolor{red}{\textbf{T2VEval (RQ\&C)}} & \textcolor{red}{\textbf{0.7098}} & \textcolor{red}{\textbf{0.7252}} & \textcolor{red}{\textbf{0.5218}} & \textcolor{red}{\textbf{0.5296}} \\
    \bottomrule
\end{tabular}
}
\label{tab5}
\end{table*}

\begin{table*}[bp]
\caption{Ablation study on the usage of technical quality branch, realness branch and alignment branch in T2VEval. \textbf{\textcolor{red}{Red}}: the best}
\centering
\scalebox{1.0}{
    \begin{tabular}{ccccccc}
        \toprule
        \textbf{Technical Quality Branch} & \textbf{Realness Branch} & \textbf{Alignment Branch} & \textbf{PLCC $\uparrow$} & \textbf{SROCC $\uparrow$} & \textbf{KROCC $\uparrow$} & \textbf{RMSE $\downarrow$} \\
        \midrule
        $\times$ & \checkmark & \checkmark & 0.6831 & 0.6829 & 0.4922 & 0.5631\\
        \checkmark & $\times$ & \checkmark & 0.6572 & 0.6713 & 0.4845 & 0.5911 \\
        \checkmark & \checkmark & $\times$ & 0.6402 & 0.6211 & 0.4475 & 0.6055 \\
        \checkmark & \checkmark & \checkmark & \textcolor{red}{\textbf{0.7098}} & \textcolor{red}{\textbf{0.7252}} & \textcolor{red}{\textbf{0.5218}} & \textcolor{red}{\textbf{0.5296}} \\
        \bottomrule
    \end{tabular}
}
\label{tab6}
\end{table*}

\subsubsection{\textbf{Ablation Experiment of Realness Branch Pre-train Strategy}}

When designing the realness evaluation branch, we introduced the pre-training task of AI-generated video detection, instead of simply using ImageNet pre-trained weights or random initialization. We replaced the pre-trained weights of Conv3D with ImageNet weights (denoted as T2VEval-ImageNet) and random initialization (denoted as T2VEval-Random) respectively, and fine-tuned them in the same training environment. The experimental results are shown in Table \ref{tab5}. In addition, we further used Grad-CAM to analyze the impact of different pre-training strategies on the model's focus area, as shown in Figure \ref{fig11}. The figure demonstrates the activation maps of the 3D network loaded with different pre-trained weights.

As shown in Figure \ref{fig11}, we present Grad-CAM heatmaps from the same layer of a 3D network for three initialization methods: random initialization, ImageNet initialization, and AI-generated video detection-based initialization. The figure illustrates results for two samples from the test set. Each sample is presented in four rows:
\begin{itemize}
\item The first row shows the original video frame and its corresponding MOS score.
\item The second row displays the heatmap, the predicted score, and the error relative to the MOS score obtained by using the initialization method proposed in this paper.
\item The third and fourth rows show the results for ImageNet initialization and random initialization, respectively.
\end{itemize}

In the first sample, the character's legs are missing in the video. Ideally, the model should focus on the areas around the legs. The heatmap reveals that our proposed initialization method effectively guides the model to focus on the lower leg's area, whereas the heatmaps from ImageNet and random initialization methods appear more disorganized. Furthermore, the prediction error demonstrates that our initialization method achieves the lowest error.

In the second sample, the legs of the horses intermittently leave the ground, disappearing and reappearing abnormally over time. The visualization results show that our initialization method produces heatmaps with significantly lower entropy than the other methods, with a sharper focus on the leg's areas in the frame. This highlights the effectiveness of our approach in guiding the model to concentrate on critical regions and achieve more accurate predictions.

This fully demonstrates the effectiveness of introducing valid and highly relevant prior information in the model training phase, which we will further explore in future work.

\subsubsection{\textbf{Ablation Experiment of The Effectiveness of Each Branch and Training Strategy}}

In this section, we performed ablation experiments to evaluate the effectiveness of each branch and the divide-and-conquer training strategy employed. Specifically, we removed each of the three branches of the model and conducted fine-tuning tests on the T2VEval-Bench dataset. The dataset partitioning method remains consistent with the previous experiments. The results, presented in Table \ref{tab6}, demonstrate that removing any branch leads to varying degrees of performance degradation, confirming the importance and effectiveness of the three dimensions assessed in the T2V quality evaluation task. Notably, removing the technical quality branch has the smallest impact on model performance, suggesting that when evaluating text-to-video generation systems, focusing solely on technical quality may prove insufficient. A comprehensive assessment framework should additionally prioritize \textbf{"content quality evaluation"}, encompassing two critical dimensions we proposed: (1) visual realness and (2) text-video alignment.

Additionally, we evaluated the impact of the divide-and-conquer training strategy we proposed by comparing it with a single-stage training approach where the entire model is trained at once using the overall impression MOS scores (denoted as T2VEval-Once). The results, shown in Table \ref{tab5}, indicate that our divide-and-conquer training strategy outperforms the single-stage approach. This finding highlights that the divide-and-conquer strategy effectively facilitates cooperation between branches, leading to improved overall performance.

\subsubsection{\textbf{Ablation Experiment of Scoring Prediction Head}}

In T2Eval, we leverage the strong capabilities of a Large Language Model (LLM) for quality regression analysis. To benchmark the LLM's performance, we compare it against commonly used linear and non-linear regression methods.
\begin{itemize}
    \item For linear regression, we employ a two-layer fully connected network: the first layer comprises 64 neurons, followed by a second layer with a single neuron. Denoted as \textbf{T2VEval-Linear}.
    \item For non-linear regression, we utilize a structure with two 1D convolutional blocks (kernel size=1): the first block has 64 channels, and the second block outputs a single channel. Denoted as \textbf{T2VEval-Conv1d.}
\end{itemize}

The results in Table \ref{tab5} show that using LLM as the score prediction head has a significant impact on improving the prediction accuracy of the model, which is closely related to the powerful reasoning ability of LLM.

\section{Conclusion}
Suffering from the infringement of the patent right of cartoon characters, we aim to propose an effective scheme to reduce the illegal use of cartoon characters. Considering the demands in patent examination and other application scenarios, we built our cartoon character dataset CCDaS with sufficient annotations, which both annotated the face and the whole-body of the cartoon characters. As for our proposed CCD algorithm MP-YOLO, we not only maintained the real-time detection speed and small parameter size of YOLO, but achieved a high detection result for our CCD task. As a result, our approach is competitive with the state-of-the-art and can be effectively used in the patent examination process or other application scenarios. In the future, we would like to explore other CCD methods and continuously improve the accuracy of CCD. \par

\section*{Acknowledgments}
This paper is supported by "the China National Science and Technology Major Project (2022ZD0116306)".\par


\bibliographystyle{cas-model2-names}

\bibliography{cas-refs}

\begin{thebibliography}{70}
\expandafter\ifx\csname natexlab\endcsname\relax\def\natexlab#1{#1}\fi
\providecommand{\url}[1]{\texttt{#1}}
\providecommand{\href}[2]{#2}
\providecommand{\path}[1]{#1}
\providecommand{\DOIprefix}{doi:}
\providecommand{\ArXivprefix}{arXiv:}
\providecommand{\URLprefix}{URL: }
\providecommand{\Pubmedprefix}{pmid:}
\providecommand{\doi}[1]{\href{http://dx.doi.org/#1}{\path{#1}}}
\providecommand{\Pubmed}[1]{\href{pmid:#1}{\path{#1}}}
\providecommand{\bibinfo}[2]{#2}
\ifx\xfnm\relax \def\xfnm[#1]{\unskip,\space#1}\fi
\bibitem[{Blattmann et~al.(2023a)Blattmann, Dockhorn, Kulal, Mendelevitch, Kilian, Lorenz, Levi, English, Voleti, Letts et~al.}]{blattmann2023stable}
\bibinfo{author}{Blattmann, A.}, \bibinfo{author}{Dockhorn, T.}, \bibinfo{author}{Kulal, S.}, \bibinfo{author}{Mendelevitch, D.}, \bibinfo{author}{Kilian, M.}, \bibinfo{author}{Lorenz, D.}, \bibinfo{author}{Levi, Y.}, \bibinfo{author}{English, Z.}, \bibinfo{author}{Voleti, V.}, \bibinfo{author}{Letts, A.}, et~al., \bibinfo{year}{2023}a.
\newblock \bibinfo{title}{Stable video diffusion: Scaling latent video diffusion models to large datasets}.
\newblock \bibinfo{journal}{arXiv preprint arXiv:2311.15127} .
\bibitem[{Blattmann et~al.(2023b)Blattmann, Rombach, Ling, Dockhorn, Kim, Fidler and Kreis}]{blattmann2023align}
\bibinfo{author}{Blattmann, A.}, \bibinfo{author}{Rombach, R.}, \bibinfo{author}{Ling, H.}, \bibinfo{author}{Dockhorn, T.}, \bibinfo{author}{Kim, S.W.}, \bibinfo{author}{Fidler, S.}, \bibinfo{author}{Kreis, K.}, \bibinfo{year}{2023}b.
\newblock \bibinfo{title}{Align your latents: High-resolution video synthesis with latent diffusion models}, in: \bibinfo{booktitle}{Proceedings of the IEEE/CVF Conference on Computer Vision and Pattern Recognition}, pp. \bibinfo{pages}{22563--22575}.
\bibitem[{Brooks et~al.(2024)Brooks, Peebles, Holmes, DePue, Guo, Jing, Schnurr, Taylor, Luhman, Luhman, Ng, Wang and Ramesh}]{videoworldsimulators2024}
\bibinfo{author}{Brooks, T.}, \bibinfo{author}{Peebles, B.}, \bibinfo{author}{Holmes, C.}, \bibinfo{author}{DePue, W.}, \bibinfo{author}{Guo, Y.}, \bibinfo{author}{Jing, L.}, \bibinfo{author}{Schnurr, D.}, \bibinfo{author}{Taylor, J.}, \bibinfo{author}{Luhman, T.}, \bibinfo{author}{Luhman, E.}, \bibinfo{author}{Ng, C.}, \bibinfo{author}{Wang, R.}, \bibinfo{author}{Ramesh, A.}, \bibinfo{year}{2024}.
\newblock \bibinfo{title}{Video generation models as world simulators} \URLprefix \url{https://openai.com/research/video-generation-models-as-world-simulators}.
\bibitem[{BT.500(2023)}]{bt2019500}
\bibinfo{author}{BT.500, I.R.}, \bibinfo{year}{2023}.
\newblock \bibinfo{title}{Methodologies for the subjective assessment of the quality of television images}.
\newblock \bibinfo{journal}{ITU-R BT.500} .
\bibitem[{Chen et~al.(2024a)Chen, Hong, Huang, Xu, Gu, Li, Lan, Zhu, Zhang, Wang et~al.}]{chen2024demamba}
\bibinfo{author}{Chen, H.}, \bibinfo{author}{Hong, Y.}, \bibinfo{author}{Huang, Z.}, \bibinfo{author}{Xu, Z.}, \bibinfo{author}{Gu, Z.}, \bibinfo{author}{Li, Y.}, \bibinfo{author}{Lan, J.}, \bibinfo{author}{Zhu, H.}, \bibinfo{author}{Zhang, J.}, \bibinfo{author}{Wang, W.}, et~al., \bibinfo{year}{2024}a.
\newblock \bibinfo{title}{Demamba: Ai-generated video detection on million-scale genvideo benchmark}.
\newblock \bibinfo{journal}{arXiv preprint arXiv:2405.19707} .
\bibitem[{Chen et~al.(2024b)Chen, Sun, Tian, Jia, Zhang, Wang, Huang, Min, Zhai and Zhang}]{chen2024gaia}
\bibinfo{author}{Chen, Z.}, \bibinfo{author}{Sun, W.}, \bibinfo{author}{Tian, Y.}, \bibinfo{author}{Jia, J.}, \bibinfo{author}{Zhang, Z.}, \bibinfo{author}{Wang, J.}, \bibinfo{author}{Huang, R.}, \bibinfo{author}{Min, X.}, \bibinfo{author}{Zhai, G.}, \bibinfo{author}{Zhang, W.}, \bibinfo{year}{2024}b.
\newblock \bibinfo{title}{Gaia: Rethinking action quality assessment for ai-generated videos}.
\newblock \bibinfo{journal}{arXiv preprint arXiv:2406.06087} .
\bibitem[{Chivileva et~al.(2023)Chivileva, Lynch, Ward and Smeaton}]{chivileva2023measuring}
\bibinfo{author}{Chivileva, I.}, \bibinfo{author}{Lynch, P.}, \bibinfo{author}{Ward, T.E.}, \bibinfo{author}{Smeaton, A.F.}, \bibinfo{year}{2023}.
\newblock \bibinfo{title}{Measuring the quality of text-to-video model outputs: Metrics and dataset}.
\newblock \bibinfo{journal}{arXiv preprint arXiv:2309.08009} .
\bibitem[{Devlin(2018)}]{devlin2018bert}
\bibinfo{author}{Devlin, J.}, \bibinfo{year}{2018}.
\newblock \bibinfo{title}{Bert: Pre-training of deep bidirectional transformers for language understanding}.
\newblock \bibinfo{journal}{arXiv preprint arXiv:1810.04805} .
\bibitem[{Feng et~al.(2024)Feng, Li, Saxon, Fu, Chen and Wang}]{feng2024tc}
\bibinfo{author}{Feng, W.}, \bibinfo{author}{Li, J.}, \bibinfo{author}{Saxon, M.}, \bibinfo{author}{Fu, T.j.}, \bibinfo{author}{Chen, W.}, \bibinfo{author}{Wang, W.Y.}, \bibinfo{year}{2024}.
\newblock \bibinfo{title}{Tc-bench: Benchmarking temporal compositionality in text-to-video and image-to-video generation}.
\newblock \bibinfo{journal}{arXiv preprint arXiv:2406.08656} .
\bibitem[{Gu et~al.(2021)Gu, Han, Liu and Huang}]{gu2021ppt}
\bibinfo{author}{Gu, Y.}, \bibinfo{author}{Han, X.}, \bibinfo{author}{Liu, Z.}, \bibinfo{author}{Huang, M.}, \bibinfo{year}{2021}.
\newblock \bibinfo{title}{Ppt: Pre-trained prompt tuning for few-shot learning}.
\newblock \bibinfo{journal}{arXiv preprint arXiv:2109.04332} .
\bibitem[{He et~al.(2022)He, Yang, Zhang, Shan and Chen}]{he2022latent}
\bibinfo{author}{He, Y.}, \bibinfo{author}{Yang, T.}, \bibinfo{author}{Zhang, Y.}, \bibinfo{author}{Shan, Y.}, \bibinfo{author}{Chen, Q.}, \bibinfo{year}{2022}.
\newblock \bibinfo{title}{Latent video diffusion models for high-fidelity long video generation}.
\newblock \bibinfo{journal}{arXiv preprint arXiv:2211.13221} .
\bibitem[{Ho et~al.(2022)Ho, Chan, Saharia, Whang, Gao, Gritsenko, Kingma, Poole, Norouzi, Fleet et~al.}]{ho2022imagen}
\bibinfo{author}{Ho, J.}, \bibinfo{author}{Chan, W.}, \bibinfo{author}{Saharia, C.}, \bibinfo{author}{Whang, J.}, \bibinfo{author}{Gao, R.}, \bibinfo{author}{Gritsenko, A.}, \bibinfo{author}{Kingma, D.P.}, \bibinfo{author}{Poole, B.}, \bibinfo{author}{Norouzi, M.}, \bibinfo{author}{Fleet, D.J.}, et~al., \bibinfo{year}{2022}.
\newblock \bibinfo{title}{Imagen video: High definition video generation with diffusion models}.
\newblock \bibinfo{journal}{arXiv preprint arXiv:2210.02303} .
\bibitem[{Hong et~al.(2022)Hong, Ding, Zheng, Liu and Tang}]{hong2022cogvideo}
\bibinfo{author}{Hong, W.}, \bibinfo{author}{Ding, M.}, \bibinfo{author}{Zheng, W.}, \bibinfo{author}{Liu, X.}, \bibinfo{author}{Tang, J.}, \bibinfo{year}{2022}.
\newblock \bibinfo{title}{Cogvideo: Large-scale pretraining for text-to-video generation via transformers}.
\newblock \bibinfo{journal}{arXiv preprint arXiv:2205.15868} .
\bibitem[{Hosu et~al.(2017a)Hosu, Hahn, Jenadeleh, Lin, Men, Szir{\'a}nyi, Li and Saupe}]{konvid1k}
\bibinfo{author}{Hosu, V.}, \bibinfo{author}{Hahn, F.}, \bibinfo{author}{Jenadeleh, M.}, \bibinfo{author}{Lin, H.}, \bibinfo{author}{Men, H.}, \bibinfo{author}{Szir{\'a}nyi, T.}, \bibinfo{author}{Li, S.}, \bibinfo{author}{Saupe, D.}, \bibinfo{year}{2017}a.
\newblock \bibinfo{title}{The konstanz natural video database}.
\newblock \URLprefix \url{http://database.mmsp-kn.de}.
\bibitem[{Hosu et~al.(2017b)Hosu, Hahn, Jenadeleh, Lin, Men, Szir{\'a}nyi, Li and Saupe}]{hosu2017konstanz}
\bibinfo{author}{Hosu, V.}, \bibinfo{author}{Hahn, F.}, \bibinfo{author}{Jenadeleh, M.}, \bibinfo{author}{Lin, H.}, \bibinfo{author}{Men, H.}, \bibinfo{author}{Szir{\'a}nyi, T.}, \bibinfo{author}{Li, S.}, \bibinfo{author}{Saupe, D.}, \bibinfo{year}{2017}b.
\newblock \bibinfo{title}{The konstanz natural video database (konvid-1k)}, in: \bibinfo{booktitle}{2017 Ninth International Conference on Quality of Multimedia Experience (QoMEX)}, \bibinfo{organization}{IEEE}. pp. \bibinfo{pages}{1--6}.
\bibitem[{Huang et~al.(2024)Huang, He, Yu, Zhang, Si, Jiang, Zhang, Wu, Jin, Chanpaisit et~al.}]{huang2024vbench}
\bibinfo{author}{Huang, Z.}, \bibinfo{author}{He, Y.}, \bibinfo{author}{Yu, J.}, \bibinfo{author}{Zhang, F.}, \bibinfo{author}{Si, C.}, \bibinfo{author}{Jiang, Y.}, \bibinfo{author}{Zhang, Y.}, \bibinfo{author}{Wu, T.}, \bibinfo{author}{Jin, Q.}, \bibinfo{author}{Chanpaisit, N.}, et~al., \bibinfo{year}{2024}.
\newblock \bibinfo{title}{Vbench: Comprehensive benchmark suite for video generative models}, in: \bibinfo{booktitle}{Proceedings of the IEEE/CVF Conference on Computer Vision and Pattern Recognition}, pp. \bibinfo{pages}{21807--21818}.
\bibitem[{Ji et~al.(2024)Ji, Xiao, Tai and Huo}]{ji2024t2vbench}
\bibinfo{author}{Ji, P.}, \bibinfo{author}{Xiao, C.}, \bibinfo{author}{Tai, H.}, \bibinfo{author}{Huo, M.}, \bibinfo{year}{2024}.
\newblock \bibinfo{title}{T2vbench: Benchmarking temporal dynamics for text-to-video generation}, in: \bibinfo{booktitle}{Proceedings of the IEEE/CVF Conference on Computer Vision and Pattern Recognition}, pp. \bibinfo{pages}{5325--5335}.
\bibitem[{Kay et~al.(2017)Kay, Carreira, Simonyan, Zhang, Hillier, Vijayanarasimhan, Viola, Green, Back, Natsev et~al.}]{kay2017kinetics}
\bibinfo{author}{Kay, W.}, \bibinfo{author}{Carreira, J.}, \bibinfo{author}{Simonyan, K.}, \bibinfo{author}{Zhang, B.}, \bibinfo{author}{Hillier, C.}, \bibinfo{author}{Vijayanarasimhan, S.}, \bibinfo{author}{Viola, F.}, \bibinfo{author}{Green, T.}, \bibinfo{author}{Back, T.}, \bibinfo{author}{Natsev, P.}, et~al., \bibinfo{year}{2017}.
\newblock \bibinfo{title}{The kinetics human action video dataset}.
\newblock \bibinfo{journal}{arXiv preprint arXiv:1705.06950} .
\bibitem[{Khachatryan et~al.(2023)Khachatryan, Movsisyan, Tadevosyan, Henschel, Wang, Navasardyan and Shi}]{khachatryan2023text2video}
\bibinfo{author}{Khachatryan, L.}, \bibinfo{author}{Movsisyan, A.}, \bibinfo{author}{Tadevosyan, V.}, \bibinfo{author}{Henschel, R.}, \bibinfo{author}{Wang, Z.}, \bibinfo{author}{Navasardyan, S.}, \bibinfo{author}{Shi, H.}, \bibinfo{year}{2023}.
\newblock \bibinfo{title}{Text2video-zero: Text-to-image diffusion models are zero-shot video generators}, in: \bibinfo{booktitle}{Proceedings of the IEEE/CVF International Conference on Computer Vision}, pp. \bibinfo{pages}{15954--15964}.
\bibitem[{Kou et~al.(2024)Kou, Liu, Zhang, Li, Wu, Min, Zhai and Liu}]{kou2024subjective}
\bibinfo{author}{Kou, T.}, \bibinfo{author}{Liu, X.}, \bibinfo{author}{Zhang, Z.}, \bibinfo{author}{Li, C.}, \bibinfo{author}{Wu, H.}, \bibinfo{author}{Min, X.}, \bibinfo{author}{Zhai, G.}, \bibinfo{author}{Liu, N.}, \bibinfo{year}{2024}.
\newblock \bibinfo{title}{Subjective-aligned dataset and metric for text-to-video quality assessment}, in: \bibinfo{booktitle}{Proceedings of the 32nd ACM International Conference on Multimedia}, pp. \bibinfo{pages}{7793--7802}.
\bibitem[{Li et~al.(2022a)Li, Zhang, Tian, Zhai and Wang}]{li2022blindly}
\bibinfo{author}{Li, B.}, \bibinfo{author}{Zhang, W.}, \bibinfo{author}{Tian, M.}, \bibinfo{author}{Zhai, G.}, \bibinfo{author}{Wang, X.}, \bibinfo{year}{2022}a.
\newblock \bibinfo{title}{Blindly assess quality of in-the-wild videos via quality-aware pre-training and motion perception}.
\newblock \bibinfo{journal}{IEEE Transactions on Circuits and Systems for Video Technology} \bibinfo{volume}{32}, \bibinfo{pages}{5944--5958}.
\bibitem[{Li et~al.(2022b)Li, Li, Xiong and Hoi}]{li2022blip}
\bibinfo{author}{Li, J.}, \bibinfo{author}{Li, D.}, \bibinfo{author}{Xiong, C.}, \bibinfo{author}{Hoi, S.}, \bibinfo{year}{2022}b.
\newblock \bibinfo{title}{Blip: Bootstrapping language-image pre-training for unified vision-language understanding and generation}, in: \bibinfo{booktitle}{International conference on machine learning}, \bibinfo{organization}{PMLR}. pp. \bibinfo{pages}{12888--12900}.
\bibitem[{Liao et~al.(2024)Liao, Lu, Zhang, Wan, Wang, Zhao, Zuo, Ye and Wang}]{liao2024evaluation}
\bibinfo{author}{Liao, M.}, \bibinfo{author}{Lu, H.}, \bibinfo{author}{Zhang, X.}, \bibinfo{author}{Wan, F.}, \bibinfo{author}{Wang, T.}, \bibinfo{author}{Zhao, Y.}, \bibinfo{author}{Zuo, W.}, \bibinfo{author}{Ye, Q.}, \bibinfo{author}{Wang, J.}, \bibinfo{year}{2024}.
\newblock \bibinfo{title}{Evaluation of text-to-video generation models: A dynamics perspective}.
\newblock \bibinfo{journal}{arXiv preprint arXiv:2407.01094} .
\bibitem[{Liu et~al.(2024a)Liu, Cun, Liu, Wang, Zhang, Chen, Liu, Zeng, Chan and Shan}]{liu2024evalcrafter}
\bibinfo{author}{Liu, Y.}, \bibinfo{author}{Cun, X.}, \bibinfo{author}{Liu, X.}, \bibinfo{author}{Wang, X.}, \bibinfo{author}{Zhang, Y.}, \bibinfo{author}{Chen, H.}, \bibinfo{author}{Liu, Y.}, \bibinfo{author}{Zeng, T.}, \bibinfo{author}{Chan, R.}, \bibinfo{author}{Shan, Y.}, \bibinfo{year}{2024}a.
\newblock \bibinfo{title}{Evalcrafter: Benchmarking and evaluating large video generation models}, in: \bibinfo{booktitle}{Proceedings of the IEEE/CVF Conference on Computer Vision and Pattern Recognition}, pp. \bibinfo{pages}{22139--22149}.
\bibitem[{Liu et~al.(2024b)Liu, Li, Ren, Gao, Li, Chen, Sun and Hou}]{liu2024fetv}
\bibinfo{author}{Liu, Y.}, \bibinfo{author}{Li, L.}, \bibinfo{author}{Ren, S.}, \bibinfo{author}{Gao, R.}, \bibinfo{author}{Li, S.}, \bibinfo{author}{Chen, S.}, \bibinfo{author}{Sun, X.}, \bibinfo{author}{Hou, L.}, \bibinfo{year}{2024}b.
\newblock \bibinfo{title}{Fetv: A benchmark for fine-grained evaluation of open-domain text-to-video generation}.
\newblock \bibinfo{journal}{Advances in Neural Information Processing Systems} \bibinfo{volume}{36}.
\bibitem[{Liu et~al.(2022a)Liu, Mao, Wu, Feichtenhofer, Darrell and Xie}]{liu2022convnet}
\bibinfo{author}{Liu, Z.}, \bibinfo{author}{Mao, H.}, \bibinfo{author}{Wu, C.Y.}, \bibinfo{author}{Feichtenhofer, C.}, \bibinfo{author}{Darrell, T.}, \bibinfo{author}{Xie, S.}, \bibinfo{year}{2022}a.
\newblock \bibinfo{title}{A convnet for the 2020s}, in: \bibinfo{booktitle}{Proceedings of the IEEE/CVF conference on computer vision and pattern recognition}, pp. \bibinfo{pages}{11976--11986}.
\bibitem[{Liu et~al.(2022b)Liu, Ning, Cao, Wei, Zhang, Lin and Hu}]{liu2022video}
\bibinfo{author}{Liu, Z.}, \bibinfo{author}{Ning, J.}, \bibinfo{author}{Cao, Y.}, \bibinfo{author}{Wei, Y.}, \bibinfo{author}{Zhang, Z.}, \bibinfo{author}{Lin, S.}, \bibinfo{author}{Hu, H.}, \bibinfo{year}{2022}b.
\newblock \bibinfo{title}{Video swin transformer}, in: \bibinfo{booktitle}{Proceedings of the IEEE/CVF conference on computer vision and pattern recognition}, pp. \bibinfo{pages}{3202--3211}.
\bibitem[{Lu et~al.(2024)Lu, Li, Li, Yu, Guan, Wang, Liao, Ye and Chen}]{lu2024aigc}
\bibinfo{author}{Lu, Y.}, \bibinfo{author}{Li, X.}, \bibinfo{author}{Li, B.}, \bibinfo{author}{Yu, Z.}, \bibinfo{author}{Guan, F.}, \bibinfo{author}{Wang, X.}, \bibinfo{author}{Liao, R.}, \bibinfo{author}{Ye, Y.}, \bibinfo{author}{Chen, Z.}, \bibinfo{year}{2024}.
\newblock \bibinfo{title}{Aigc-vqa: A holistic perception metric for aigc video quality assessment}, in: \bibinfo{booktitle}{Proceedings of the IEEE/CVF Conference on Computer Vision and Pattern Recognition}, pp. \bibinfo{pages}{6384--6394}.
\bibitem[{Luo et~al.(2023)Luo, Chen, Zhang, Huang, Wang, Shen, Zhao, Zhou and Tan}]{luo2023videofusion}
\bibinfo{author}{Luo, Z.}, \bibinfo{author}{Chen, D.}, \bibinfo{author}{Zhang, Y.}, \bibinfo{author}{Huang, Y.}, \bibinfo{author}{Wang, L.}, \bibinfo{author}{Shen, Y.}, \bibinfo{author}{Zhao, D.}, \bibinfo{author}{Zhou, J.}, \bibinfo{author}{Tan, T.}, \bibinfo{year}{2023}.
\newblock \bibinfo{title}{Videofusion: Decomposed diffusion models for high-quality video generation}.
\newblock \bibinfo{journal}{arXiv preprint arXiv:2303.08320} .
\bibitem[{Ma et~al.(2019)Ma, Yu, Wu and Wang}]{ma2019paddlepaddle}
\bibinfo{author}{Ma, Y.}, \bibinfo{author}{Yu, D.}, \bibinfo{author}{Wu, T.}, \bibinfo{author}{Wang, H.}, \bibinfo{year}{2019}.
\newblock \bibinfo{title}{Paddlepaddle: An open-source deep learning platform from industrial practice}.
\newblock \bibinfo{journal}{Frontiers of Data and Domputing} \bibinfo{volume}{1}, \bibinfo{pages}{105--115}.
\bibitem[{Miao et~al.(2024)Miao, Zhu, Dong, Yu, Zhu and Gao}]{miao2024t2vsafetybench}
\bibinfo{author}{Miao, Y.}, \bibinfo{author}{Zhu, Y.}, \bibinfo{author}{Dong, Y.}, \bibinfo{author}{Yu, L.}, \bibinfo{author}{Zhu, J.}, \bibinfo{author}{Gao, X.S.}, \bibinfo{year}{2024}.
\newblock \bibinfo{title}{T2vsafetybench: Evaluating the safety of text-to-video generative models}.
\newblock \bibinfo{journal}{arXiv preprint arXiv:2407.05965} .
\bibitem[{Min et~al.(2024a)Min, Duan, Sun, Zhu and Zhai}]{min2024perceptualvideoqualityassessment}
\bibinfo{author}{Min, X.}, \bibinfo{author}{Duan, H.}, \bibinfo{author}{Sun, W.}, \bibinfo{author}{Zhu, Y.}, \bibinfo{author}{Zhai, G.}, \bibinfo{year}{2024}a.
\newblock \bibinfo{title}{Perceptual video quality assessment: A survey}.
\newblock \URLprefix \url{https://arxiv.org/abs/2402.03413}, \href{http://arxiv.org/abs/2402.03413}{\tt arXiv:2402.03413}.
\bibitem[{Min et~al.(2024b)Min, Gao, Cao, Zhai, Zhang, Sun and Chen}]{min2024exploring}
\bibinfo{author}{Min, X.}, \bibinfo{author}{Gao, Y.}, \bibinfo{author}{Cao, Y.}, \bibinfo{author}{Zhai, G.}, \bibinfo{author}{Zhang, W.}, \bibinfo{author}{Sun, H.}, \bibinfo{author}{Chen, C.W.}, \bibinfo{year}{2024}b.
\newblock \bibinfo{title}{Exploring rich subjective quality information for image quality assessment in the wild}.
\newblock \bibinfo{journal}{arXiv preprint arXiv:2409.05540} .
\bibitem[{Min et~al.(2017)Min, Gu, Zhai, Liu, Yang and Chen}]{min2017blind}
\bibinfo{author}{Min, X.}, \bibinfo{author}{Gu, K.}, \bibinfo{author}{Zhai, G.}, \bibinfo{author}{Liu, J.}, \bibinfo{author}{Yang, X.}, \bibinfo{author}{Chen, C.W.}, \bibinfo{year}{2017}.
\newblock \bibinfo{title}{Blind quality assessment based on pseudo-reference image}.
\newblock \bibinfo{journal}{IEEE Transactions on Multimedia} \bibinfo{volume}{20}, \bibinfo{pages}{2049--2062}.
\bibitem[{Min et~al.(2021)Min, Gu, Zhai, Yang, Zhang, Le~Callet and Chen}]{10.1145/3470970}
\bibinfo{author}{Min, X.}, \bibinfo{author}{Gu, K.}, \bibinfo{author}{Zhai, G.}, \bibinfo{author}{Yang, X.}, \bibinfo{author}{Zhang, W.}, \bibinfo{author}{Le~Callet, P.}, \bibinfo{author}{Chen, C.W.}, \bibinfo{year}{2021}.
\newblock \bibinfo{title}{Screen content quality assessment: Overview, benchmark, and beyond}.
\newblock \bibinfo{journal}{ACM Comput. Surv.} \bibinfo{volume}{54}.
\newblock \URLprefix \url{https://doi.org/10.1145/3470970}, \DOIprefix\doi{10.1145/3470970}.
\bibitem[{Min et~al.(2018)Min, Zhai, Gu, Liu and Yang}]{min2018blind}
\bibinfo{author}{Min, X.}, \bibinfo{author}{Zhai, G.}, \bibinfo{author}{Gu, K.}, \bibinfo{author}{Liu, Y.}, \bibinfo{author}{Yang, X.}, \bibinfo{year}{2018}.
\newblock \bibinfo{title}{Blind image quality estimation via distortion aggravation}.
\newblock \bibinfo{journal}{IEEE Transactions on Broadcasting} \bibinfo{volume}{64}, \bibinfo{pages}{508--517}.
\bibitem[{Radford et~al.(2021)Radford, Kim, Hallacy, Ramesh, Goh, Agarwal, Sastry, Askell, Mishkin, Clark et~al.}]{radford2021learning}
\bibinfo{author}{Radford, A.}, \bibinfo{author}{Kim, J.W.}, \bibinfo{author}{Hallacy, C.}, \bibinfo{author}{Ramesh, A.}, \bibinfo{author}{Goh, G.}, \bibinfo{author}{Agarwal, S.}, \bibinfo{author}{Sastry, G.}, \bibinfo{author}{Askell, A.}, \bibinfo{author}{Mishkin, P.}, \bibinfo{author}{Clark, J.}, et~al., \bibinfo{year}{2021}.
\newblock \bibinfo{title}{Learning transferable visual models from natural language supervision}, in: \bibinfo{booktitle}{International conference on machine learning}, \bibinfo{organization}{PMLR}. pp. \bibinfo{pages}{8748--8763}.
\bibitem[{Singer et~al.(2022)Singer, Polyak, Hayes, Yin, An, Zhang, Hu, Yang, Ashual, Gafni et~al.}]{singer2022make}
\bibinfo{author}{Singer, U.}, \bibinfo{author}{Polyak, A.}, \bibinfo{author}{Hayes, T.}, \bibinfo{author}{Yin, X.}, \bibinfo{author}{An, J.}, \bibinfo{author}{Zhang, S.}, \bibinfo{author}{Hu, Q.}, \bibinfo{author}{Yang, H.}, \bibinfo{author}{Ashual, O.}, \bibinfo{author}{Gafni, O.}, et~al., \bibinfo{year}{2022}.
\newblock \bibinfo{title}{Make-a-video: Text-to-video generation without text-video data}.
\newblock \bibinfo{journal}{arXiv preprint arXiv:2209.14792} .
\bibitem[{Sinno and Bovik(2018)}]{sinno2018large}
\bibinfo{author}{Sinno, Z.}, \bibinfo{author}{Bovik, A.C.}, \bibinfo{year}{2018}.
\newblock \bibinfo{title}{Large-scale study of perceptual video quality}.
\newblock \bibinfo{journal}{IEEE Transactions on Image Processing} \bibinfo{volume}{28}, \bibinfo{pages}{612--627}.
\bibitem[{Sun et~al.(2024)Sun, Huang, Liu, Wu, Xu, Li and Liu}]{sun2024t2v}
\bibinfo{author}{Sun, K.}, \bibinfo{author}{Huang, K.}, \bibinfo{author}{Liu, X.}, \bibinfo{author}{Wu, Y.}, \bibinfo{author}{Xu, Z.}, \bibinfo{author}{Li, Z.}, \bibinfo{author}{Liu, X.}, \bibinfo{year}{2024}.
\newblock \bibinfo{title}{T2v-compbench: A comprehensive benchmark for compositional text-to-video generation}.
\newblock \bibinfo{journal}{arXiv preprint arXiv:2407.14505} .
\bibitem[{Sun et~al.(2025)Sun, Fu, Cao, Zhu, Zhang, Zhu, Zhang, Hu, Min and Zhai}]{sun2025empirical}
\bibinfo{author}{Sun, W.}, \bibinfo{author}{Fu, K.}, \bibinfo{author}{Cao, L.}, \bibinfo{author}{Zhu, D.}, \bibinfo{author}{Zhang, K.}, \bibinfo{author}{Zhu, Y.}, \bibinfo{author}{Zhang, Z.}, \bibinfo{author}{Hu, M.}, \bibinfo{author}{Min, X.}, \bibinfo{author}{Zhai, G.}, \bibinfo{year}{2025}.
\newblock \bibinfo{title}{An empirical study for efficient video quality assessment}, in: \bibinfo{booktitle}{Proceedings of the Computer Vision and Pattern Recognition Conference}, pp. \bibinfo{pages}{1403--1413}.
\bibitem[{Sun et~al.(2022)Sun, Min, Lu and Zhai}]{sun2022deep}
\bibinfo{author}{Sun, W.}, \bibinfo{author}{Min, X.}, \bibinfo{author}{Lu, W.}, \bibinfo{author}{Zhai, G.}, \bibinfo{year}{2022}.
\newblock \bibinfo{title}{A deep learning based no-reference quality assessment model for ugc videos}, in: \bibinfo{booktitle}{Proceedings of the 30th ACM International Conference on Multimedia}, pp. \bibinfo{pages}{856--865}.
\bibitem[{Teed and Deng(2020)}]{teed2020raft}
\bibinfo{author}{Teed, Z.}, \bibinfo{author}{Deng, J.}, \bibinfo{year}{2020}.
\newblock \bibinfo{title}{Raft: Recurrent all-pairs field transforms for optical flow}, in: \bibinfo{booktitle}{Computer Vision--ECCV 2020: 16th European Conference, Glasgow, UK, August 23--28, 2020, Proceedings, Part II 16}, \bibinfo{organization}{Springer}. pp. \bibinfo{pages}{402--419}.
\bibitem[{Touvron et~al.(2023)Touvron, Martin, Stone, Albert, Almahairi, Babaei, Bashlykov, Batra, Bhargava, Bhosale et~al.}]{touvron2023llama}
\bibinfo{author}{Touvron, H.}, \bibinfo{author}{Martin, L.}, \bibinfo{author}{Stone, K.}, \bibinfo{author}{Albert, P.}, \bibinfo{author}{Almahairi, A.}, \bibinfo{author}{Babaei, Y.}, \bibinfo{author}{Bashlykov, N.}, \bibinfo{author}{Batra, S.}, \bibinfo{author}{Bhargava, P.}, \bibinfo{author}{Bhosale, S.}, et~al., \bibinfo{year}{2023}.
\newblock \bibinfo{title}{Llama 2: Open foundation and fine-tuned chat models}.
\newblock \bibinfo{journal}{arXiv preprint arXiv:2307.09288} .
\bibitem[{Villegas et~al.(2022)Villegas, Babaeizadeh, Kindermans, Moraldo, Zhang, Saffar, Castro, Kunze and Erhan}]{villegas2022phenaki}
\bibinfo{author}{Villegas, R.}, \bibinfo{author}{Babaeizadeh, M.}, \bibinfo{author}{Kindermans, P.J.}, \bibinfo{author}{Moraldo, H.}, \bibinfo{author}{Zhang, H.}, \bibinfo{author}{Saffar, M.T.}, \bibinfo{author}{Castro, S.}, \bibinfo{author}{Kunze, J.}, \bibinfo{author}{Erhan, D.}, \bibinfo{year}{2022}.
\newblock \bibinfo{title}{Phenaki: Variable length video generation from open domain textual descriptions}, in: \bibinfo{booktitle}{International Conference on Learning Representations}.
\bibitem[{Wang et~al.(2025a)Wang, Duan, Zhai and Min}]{wang2025qualityassessmentaigenerated}
\bibinfo{author}{Wang, J.}, \bibinfo{author}{Duan, H.}, \bibinfo{author}{Zhai, G.}, \bibinfo{author}{Min, X.}, \bibinfo{year}{2025}a.
\newblock \bibinfo{title}{Quality assessment for ai generated images with instruction tuning}.
\newblock \URLprefix \url{https://arxiv.org/abs/2405.07346}, \href{http://arxiv.org/abs/2405.07346}{\tt arXiv:2405.07346}.
\bibitem[{Wang et~al.(2024a)Wang, Duan, Zhai, Wang and Min}]{wang2024aigvassessorbenchmarkingevaluatingperceptual}
\bibinfo{author}{Wang, J.}, \bibinfo{author}{Duan, H.}, \bibinfo{author}{Zhai, G.}, \bibinfo{author}{Wang, J.}, \bibinfo{author}{Min, X.}, \bibinfo{year}{2024}a.
\newblock \bibinfo{title}{Aigv-assessor: Benchmarking and evaluating the perceptual quality of text-to-video generation with lmm}.
\newblock \URLprefix \url{https://arxiv.org/abs/2411.17221}, \href{http://arxiv.org/abs/2411.17221}{\tt arXiv:2411.17221}.
\bibitem[{Wang et~al.(2025b)Wang, Duan, Zhao, Wang, Zhai and Min}]{wang2025lmm4lmmbenchmarkingevaluatinglargemultimodal}
\bibinfo{author}{Wang, J.}, \bibinfo{author}{Duan, H.}, \bibinfo{author}{Zhao, Y.}, \bibinfo{author}{Wang, J.}, \bibinfo{author}{Zhai, G.}, \bibinfo{author}{Min, X.}, \bibinfo{year}{2025}b.
\newblock \bibinfo{title}{Lmm4lmm: Benchmarking and evaluating large-multimodal image generation with lmms}.
\newblock \URLprefix \url{https://arxiv.org/abs/2504.08358}, \href{http://arxiv.org/abs/2504.08358}{\tt arXiv:2504.08358}.
\bibitem[{Wang et~al.(2023)Wang, Yuan, Chen, Zhang, Wang and Zhang}]{wang2023modelscope}
\bibinfo{author}{Wang, J.}, \bibinfo{author}{Yuan, H.}, \bibinfo{author}{Chen, D.}, \bibinfo{author}{Zhang, Y.}, \bibinfo{author}{Wang, X.}, \bibinfo{author}{Zhang, S.}, \bibinfo{year}{2023}.
\newblock \bibinfo{title}{Modelscope text-to-video technical report}.
\newblock \bibinfo{journal}{arXiv preprint arXiv:2308.06571} .
\bibitem[{Wang et~al.(2024b)Wang, Zhang, Yuan, Qing, Gong, Zhang, Shen, Gao and Sang}]{wang2024recipe}
\bibinfo{author}{Wang, X.}, \bibinfo{author}{Zhang, S.}, \bibinfo{author}{Yuan, H.}, \bibinfo{author}{Qing, Z.}, \bibinfo{author}{Gong, B.}, \bibinfo{author}{Zhang, Y.}, \bibinfo{author}{Shen, Y.}, \bibinfo{author}{Gao, C.}, \bibinfo{author}{Sang, N.}, \bibinfo{year}{2024}b.
\newblock \bibinfo{title}{A recipe for scaling up text-to-video generation with text-free videos}, in: \bibinfo{booktitle}{Proceedings of the IEEE/CVF Conference on Computer Vision and Pattern Recognition}, pp. \bibinfo{pages}{6572--6582}.
\bibitem[{Wang et~al.(2019)Wang, Inguva and Adsumilli}]{wang2019youtube}
\bibinfo{author}{Wang, Y.}, \bibinfo{author}{Inguva, S.}, \bibinfo{author}{Adsumilli, B.}, \bibinfo{year}{2019}.
\newblock \bibinfo{title}{Youtube ugc dataset for video compression research}, in: \bibinfo{booktitle}{2019 IEEE 21st International Workshop on Multimedia Signal Processing (MMSP)}, \bibinfo{organization}{IEEE}. pp. \bibinfo{pages}{1--5}.
\bibitem[{Wu et~al.(2021)Wu, Huang, Zhang, Li, Ji, Yang, Sapiro and Duan}]{wu2021godiva}
\bibinfo{author}{Wu, C.}, \bibinfo{author}{Huang, L.}, \bibinfo{author}{Zhang, Q.}, \bibinfo{author}{Li, B.}, \bibinfo{author}{Ji, L.}, \bibinfo{author}{Yang, F.}, \bibinfo{author}{Sapiro, G.}, \bibinfo{author}{Duan, N.}, \bibinfo{year}{2021}.
\newblock \bibinfo{title}{Godiva: Generating open-domain videos from natural descriptions}.
\newblock \bibinfo{journal}{arXiv preprint arXiv:2104.14806} .
\bibitem[{Wu et~al.(2022a)Wu, Liang, Ji, Yang, Fang, Jiang and Duan}]{wu2022nuwa}
\bibinfo{author}{Wu, C.}, \bibinfo{author}{Liang, J.}, \bibinfo{author}{Ji, L.}, \bibinfo{author}{Yang, F.}, \bibinfo{author}{Fang, Y.}, \bibinfo{author}{Jiang, D.}, \bibinfo{author}{Duan, N.}, \bibinfo{year}{2022}a.
\newblock \bibinfo{title}{N{\"u}wa: Visual synthesis pre-training for neural visual world creation}, in: \bibinfo{booktitle}{European conference on computer vision}, \bibinfo{organization}{Springer}. pp. \bibinfo{pages}{720--736}.
\bibitem[{Wu et~al.(2022b)Wu, Chen, Hou, Liao, Wang, Sun, Yan and Lin}]{wu2022fast}
\bibinfo{author}{Wu, H.}, \bibinfo{author}{Chen, C.}, \bibinfo{author}{Hou, J.}, \bibinfo{author}{Liao, L.}, \bibinfo{author}{Wang, A.}, \bibinfo{author}{Sun, W.}, \bibinfo{author}{Yan, Q.}, \bibinfo{author}{Lin, W.}, \bibinfo{year}{2022}b.
\newblock \bibinfo{title}{Fast-vqa: Efficient end-to-end video quality assessment with fragment sampling}, in: \bibinfo{booktitle}{European conference on computer vision}, \bibinfo{organization}{Springer}. pp. \bibinfo{pages}{538--554}.
\bibitem[{Wu et~al.(2023a)Wu, Zhang, Liao, Chen, Hou, Wang, Sun, Yan and Lin}]{wu2023exploring}
\bibinfo{author}{Wu, H.}, \bibinfo{author}{Zhang, E.}, \bibinfo{author}{Liao, L.}, \bibinfo{author}{Chen, C.}, \bibinfo{author}{Hou, J.}, \bibinfo{author}{Wang, A.}, \bibinfo{author}{Sun, W.}, \bibinfo{author}{Yan, Q.}, \bibinfo{author}{Lin, W.}, \bibinfo{year}{2023}a.
\newblock \bibinfo{title}{Exploring video quality assessment on user generated contents from aesthetic and technical perspectives}, in: \bibinfo{booktitle}{Proceedings of the IEEE/CVF International Conference on Computer Vision}, pp. \bibinfo{pages}{20144--20154}.
\bibitem[{Wu et~al.(2024a)Wu, Zhang, Zhang, Chen, Liao, Wang, Xu, Li, Hou, Zhai et~al.}]{wu2024q}
\bibinfo{author}{Wu, H.}, \bibinfo{author}{Zhang, Z.}, \bibinfo{author}{Zhang, E.}, \bibinfo{author}{Chen, C.}, \bibinfo{author}{Liao, L.}, \bibinfo{author}{Wang, A.}, \bibinfo{author}{Xu, K.}, \bibinfo{author}{Li, C.}, \bibinfo{author}{Hou, J.}, \bibinfo{author}{Zhai, G.}, et~al., \bibinfo{year}{2024}a.
\newblock \bibinfo{title}{Q-instruct: Improving low-level visual abilities for multi-modality foundation models}, in: \bibinfo{booktitle}{Proceedings of the IEEE/CVF Conference on Computer Vision and Pattern Recognition}, pp. \bibinfo{pages}{25490--25500}.
\bibitem[{Wu et~al.(2023b)Wu, Zhang, Zhang, Chen, Liao, Li, Gao, Wang, Zhang, Sun et~al.}]{wu2023q}
\bibinfo{author}{Wu, H.}, \bibinfo{author}{Zhang, Z.}, \bibinfo{author}{Zhang, W.}, \bibinfo{author}{Chen, C.}, \bibinfo{author}{Liao, L.}, \bibinfo{author}{Li, C.}, \bibinfo{author}{Gao, Y.}, \bibinfo{author}{Wang, A.}, \bibinfo{author}{Zhang, E.}, \bibinfo{author}{Sun, W.}, et~al., \bibinfo{year}{2023}b.
\newblock \bibinfo{title}{Q-align: Teaching lmms for visual scoring via discrete text-defined levels}.
\newblock \bibinfo{journal}{arXiv preprint arXiv:2312.17090} .
\bibitem[{Wu et~al.(2025)Wu, Zhu, Zhang, Zhang, Chen, Liao, Li, Wang, Sun, Yan et~al.}]{wu2025towards}
\bibinfo{author}{Wu, H.}, \bibinfo{author}{Zhu, H.}, \bibinfo{author}{Zhang, Z.}, \bibinfo{author}{Zhang, E.}, \bibinfo{author}{Chen, C.}, \bibinfo{author}{Liao, L.}, \bibinfo{author}{Li, C.}, \bibinfo{author}{Wang, A.}, \bibinfo{author}{Sun, W.}, \bibinfo{author}{Yan, Q.}, et~al., \bibinfo{year}{2025}.
\newblock \bibinfo{title}{Towards open-ended visual quality comparison}, in: \bibinfo{booktitle}{European Conference on Computer Vision}, \bibinfo{organization}{Springer}. pp. \bibinfo{pages}{360--377}.
\bibitem[{Wu et~al.(2024b)Wu, Fang, Wu, Wang, Ge, Cun, Zhang, Liu, Gu, Zhao et~al.}]{wu2024towards}
\bibinfo{author}{Wu, J.Z.}, \bibinfo{author}{Fang, G.}, \bibinfo{author}{Wu, H.}, \bibinfo{author}{Wang, X.}, \bibinfo{author}{Ge, Y.}, \bibinfo{author}{Cun, X.}, \bibinfo{author}{Zhang, D.J.}, \bibinfo{author}{Liu, J.W.}, \bibinfo{author}{Gu, Y.}, \bibinfo{author}{Zhao, R.}, et~al., \bibinfo{year}{2024}b.
\newblock \bibinfo{title}{Towards a better metric for text-to-video generation}.
\newblock \bibinfo{journal}{arXiv preprint arXiv:2401.07781} .
\bibitem[{Xu et~al.(2024)Xu, Liu, Wu, Tong, Li, Ding, Tang and Dong}]{xu2024imagereward}
\bibinfo{author}{Xu, J.}, \bibinfo{author}{Liu, X.}, \bibinfo{author}{Wu, Y.}, \bibinfo{author}{Tong, Y.}, \bibinfo{author}{Li, Q.}, \bibinfo{author}{Ding, M.}, \bibinfo{author}{Tang, J.}, \bibinfo{author}{Dong, Y.}, \bibinfo{year}{2024}.
\newblock \bibinfo{title}{Imagereward: Learning and evaluating human preferences for text-to-image generation}.
\newblock \bibinfo{journal}{Advances in Neural Information Processing Systems} \bibinfo{volume}{36}.
\bibitem[{Yang et~al.(2024)Yang, Teng, Zheng, Ding, Huang, Xu, Yang, Hong, Zhang, Feng et~al.}]{yang2024cogvideox}
\bibinfo{author}{Yang, Z.}, \bibinfo{author}{Teng, J.}, \bibinfo{author}{Zheng, W.}, \bibinfo{author}{Ding, M.}, \bibinfo{author}{Huang, S.}, \bibinfo{author}{Xu, J.}, \bibinfo{author}{Yang, Y.}, \bibinfo{author}{Hong, W.}, \bibinfo{author}{Zhang, X.}, \bibinfo{author}{Feng, G.}, et~al., \bibinfo{year}{2024}.
\newblock \bibinfo{title}{Cogvideox: Text-to-video diffusion models with an expert transformer}.
\newblock \bibinfo{journal}{arXiv preprint arXiv:2408.06072} .
\bibitem[{Zhai Guang~Tao(2020)}]{zhai2020iqasurvey}
\bibinfo{author}{Zhai Guang~Tao, M.X.K.}, \bibinfo{year}{2020}.
\newblock \bibinfo{title}{Perceptual image quality assessment: a survey}.
\newblock \bibinfo{journal}{Science China Information Sciences} .
\bibitem[{Zhang et~al.(2024a)Zhang, Wu, Liu, Zhao, Ran, Gu, Gao and Shou}]{zhang2024show}
\bibinfo{author}{Zhang, D.J.}, \bibinfo{author}{Wu, J.Z.}, \bibinfo{author}{Liu, J.W.}, \bibinfo{author}{Zhao, R.}, \bibinfo{author}{Ran, L.}, \bibinfo{author}{Gu, Y.}, \bibinfo{author}{Gao, D.}, \bibinfo{author}{Shou, M.Z.}, \bibinfo{year}{2024}a.
\newblock \bibinfo{title}{Show-1: Marrying pixel and latent diffusion models for text-to-video generation}.
\newblock \bibinfo{journal}{International Journal of Computer Vision} , \bibinfo{pages}{1--15}.
\bibitem[{Zhang et~al.(2024b)Zhang, Ma, Yan, Zhang, Wang, Yang, Guo, Shao, You, Qiao et~al.}]{zhang2024rethinking}
\bibinfo{author}{Zhang, T.}, \bibinfo{author}{Ma, L.}, \bibinfo{author}{Yan, Y.}, \bibinfo{author}{Zhang, Y.}, \bibinfo{author}{Wang, K.}, \bibinfo{author}{Yang, Y.}, \bibinfo{author}{Guo, Z.}, \bibinfo{author}{Shao, W.}, \bibinfo{author}{You, Y.}, \bibinfo{author}{Qiao, Y.}, et~al., \bibinfo{year}{2024}b.
\newblock \bibinfo{title}{Rethinking human evaluation protocol for text-to-video models: Enhancing reliability, reproducibility, and practicality}.
\newblock \bibinfo{journal}{arXiv preprint arXiv:2406.08845} .
\bibitem[{Zhang et~al.(2024c)Zhang, Wang, Zhu and Xie}]{ZHANG2024102719}
\bibinfo{author}{Zhang, Y.}, \bibinfo{author}{Wang, J.}, \bibinfo{author}{Zhu, Y.}, \bibinfo{author}{Xie, R.}, \bibinfo{year}{2024}c.
\newblock \bibinfo{title}{Subjective and objective quality evaluation of ugc video after encoding and decoding}.
\newblock \bibinfo{journal}{Displays} \bibinfo{volume}{83}, \bibinfo{pages}{102719}.
\newblock \URLprefix \url{https://www.sciencedirect.com/science/article/pii/S0141938224000830}, \DOIprefix\doi{https://doi.org/10.1016/j.displa.2024.102719}.
\bibitem[{Zhang et~al.(2024d)Zhang, Li, Sun, Jia, Min, Zhang, Li, Chen, Wang, Ji et~al.}]{zhang2024benchmarking}
\bibinfo{author}{Zhang, Z.}, \bibinfo{author}{Li, X.}, \bibinfo{author}{Sun, W.}, \bibinfo{author}{Jia, J.}, \bibinfo{author}{Min, X.}, \bibinfo{author}{Zhang, Z.}, \bibinfo{author}{Li, C.}, \bibinfo{author}{Chen, Z.}, \bibinfo{author}{Wang, P.}, \bibinfo{author}{Ji, Z.}, et~al., \bibinfo{year}{2024}d.
\newblock \bibinfo{title}{Benchmarking aigc video quality assessment: A dataset and unified model}.
\newblock \bibinfo{journal}{arXiv preprint arXiv:2407.21408} .
\bibitem[{Zhang et~al.(2024e)Zhang, Wu, Ji, Li, Zhang, Sun, Liu, Min, Sun, Jui et~al.}]{zhang2024q}
\bibinfo{author}{Zhang, Z.}, \bibinfo{author}{Wu, H.}, \bibinfo{author}{Ji, Z.}, \bibinfo{author}{Li, C.}, \bibinfo{author}{Zhang, E.}, \bibinfo{author}{Sun, W.}, \bibinfo{author}{Liu, X.}, \bibinfo{author}{Min, X.}, \bibinfo{author}{Sun, F.}, \bibinfo{author}{Jui, S.}, et~al., \bibinfo{year}{2024}e.
\newblock \bibinfo{title}{Q-boost: On visual quality assessment ability of low-level multi-modality foundation models}, in: \bibinfo{booktitle}{2024 IEEE International Conference on Multimedia and Expo Workshops (ICMEW)}, \bibinfo{organization}{IEEE}. pp. \bibinfo{pages}{1--6}.
\bibitem[{Zheng et~al.(2023)Zheng, Chiang, Sheng, Zhuang, Wu, Zhuang, Lin, Li, Li, Xing et~al.}]{zheng2023judging}
\bibinfo{author}{Zheng, L.}, \bibinfo{author}{Chiang, W.L.}, \bibinfo{author}{Sheng, Y.}, \bibinfo{author}{Zhuang, S.}, \bibinfo{author}{Wu, Z.}, \bibinfo{author}{Zhuang, Y.}, \bibinfo{author}{Lin, Z.}, \bibinfo{author}{Li, Z.}, \bibinfo{author}{Li, D.}, \bibinfo{author}{Xing, E.}, et~al., \bibinfo{year}{2023}.
\newblock \bibinfo{title}{Judging llm-as-a-judge with mt-bench and chatbot arena}.
\newblock \bibinfo{journal}{Advances in Neural Information Processing Systems} \bibinfo{volume}{36}, \bibinfo{pages}{46595--46623}.
\bibitem[{Zhou et~al.(2022)Zhou, Wang, Yan, Lv, Zhu and Feng}]{zhou2022magicvideo}
\bibinfo{author}{Zhou, D.}, \bibinfo{author}{Wang, W.}, \bibinfo{author}{Yan, H.}, \bibinfo{author}{Lv, W.}, \bibinfo{author}{Zhu, Y.}, \bibinfo{author}{Feng, J.}, \bibinfo{year}{2022}.
\newblock \bibinfo{title}{Magicvideo: Efficient video generation with latent diffusion models}.
\newblock \bibinfo{journal}{arXiv preprint arXiv:2211.11018} .
\bibitem[{Zhu et~al.(2023)Zhu, Li, Sun, Min, Zhai and Yang}]{9961939}
\bibinfo{author}{Zhu, Y.}, \bibinfo{author}{Li, Y.}, \bibinfo{author}{Sun, W.}, \bibinfo{author}{Min, X.}, \bibinfo{author}{Zhai, G.}, \bibinfo{author}{Yang, X.}, \bibinfo{year}{2023}.
\newblock \bibinfo{title}{Blind image quality assessment via cross-view consistency}.
\newblock \bibinfo{journal}{IEEE Transactions on Multimedia} \bibinfo{volume}{25}, \bibinfo{pages}{7607--7620}.
\newblock \DOIprefix\doi{10.1109/TMM.2022.3224319}.

\end{thebibliography}



\end{document}